%% file: main.tex
\documentclass[times,twocolumn,final,authoryear]{elsarticle}

\usepackage{ycviu}
\usepackage{framed,multirow}

\usepackage{amssymb}
\usepackage{latexsym}

\usepackage{graphicx}
\usepackage{amsmath}
\usepackage{url}            
\usepackage{booktabs}      
\usepackage{amsfonts}        
\usepackage{array}
\usepackage{color}
\usepackage{colortbl}
\usepackage{framed}
\usepackage{bm}
\usepackage{bbm}
\usepackage{xspace}
\usepackage{enumitem}
\usepackage{xparse}
\usepackage{algorithm}
\usepackage{listings}
\usepackage{csquotes}

\usepackage{url}
\usepackage{xcolor}
\usepackage{pifont}

\newcommand{\xmark}{\ding{55}}%

\definecolor{Gray}{gray}{0.9}
\definecolor{LightCyan}{rgb}{0.88,0.95,1}
\definecolor{blond}{rgb}{0.98, 0.94, 0.75}

\def \ie {\emph{i.e.}}

\newcommand{\tit}[1]{\smallbreak\noindent\textbf{#1.}}
\newcommand{\tinytit}[1]{\noindent\textbf{#1.}}

\newcommand{\ours}{MaPeT\xspace}

\newcommand{\rev}[1]{\textcolor{black}{#1}}

\journal{Computer Vision and Image Understanding}

\begin{document}

\begin{frontmatter}

\title{Learning to Mask and Permute Visual Tokens for Vision Transformer Pre-Training}

\author[]{Lorenzo \snm{Baraldi}$^{\text{a}\hspace{0.05cm}*\hspace{0.05cm}}$}
\author[]{Roberto \snm{Amoroso}$^{\text{b}\hspace{0.05cm}*}$}
\author[2]{Marcella \snm{Cornia}}
\author[2]{Lorenzo \snm{Baraldi}}
\author[3]{Andrea \snm{Pilzer}}
\author[2,4]{Rita \snm{Cucchiara}}


\address[1]{University of Pisa, Pisa, Italy}
\address[2]{University of Modena and Reggio Emilia, Modena, Italy}
\address[3]{NVIDIA AI Technology Center, Italy}
\address[4]{IIT-CNR, Pisa, Italy}

\begin{abstract}
The use of self-supervised pre-training has emerged as a promising approach to enhance the performance of many different visual tasks. In this context, recent approaches have employed the Masked Image Modeling paradigm, which pre-trains a backbone by reconstructing visual tokens associated with randomly masked image patches. This masking approach, however, introduces noise into the input data during pre-training, leading to discrepancies that can impair performance during the fine-tuning phase. Furthermore, input masking neglects the dependencies between corrupted patches, increasing the inconsistencies observed in downstream fine-tuning tasks. To overcome these issues, we propose a new self-supervised pre-training approach, named Masked and Permuted Vision Transformer (\ours), that employs autoregressive and permuted predictions to capture intra-patch dependencies. In addition, \ours employs auxiliary positional information to reduce the disparity between the pre-training and fine-tuning phases. In our experiments, we employ a fair setting to ensure reliable and meaningful comparisons and conduct investigations on multiple visual tokenizers, including our proposed $k$-CLIP which directly employs discretized CLIP features. Our results demonstrate that \ours achieves competitive performance on ImageNet, compared to baselines and competitors under the same model setting. We release an implementation of our code and models at \small{\url{https://github.com/aimagelab/MaPeT}}.
\end{abstract}

\end{frontmatter}

\section{Introduction}
\label{sec:intro}
\input{sections/01_introduction.tex}

\section{Related Work}
\label{sec:related}
\input{sections/02_related.tex}

\section{Preliminaries}
\label{sec:preliminaries}
\input{sections/03_preliminaries.tex}

\section{Proposed Method}
\label{sec:method}
\input{sections/04_method.tex}

\section{Experiments}
\label{sec:experiments}
\input{sections/05_experiments.tex}

\section{Conclusion}
\label{sec:conclusion}
\input{sections/06_conclusion.tex}

\section*{Acknowledgments}
We acknowledge the CINECA award under the ISCRA initiative, for the availability of high-performance computing resources. This work has partially been supported by the PNRR-M4C2 (PE00000013) project “FAIR - Future Artificial Intelligence Research” and the Horizon Europe project “ELSA - European Lighthouse on Secure and Safe AI” (GA 101070617), both funded by the European Union.

\bibliographystyle{model2-names}
\bibliography{bibliography}

\clearpage
\newpage
\section*{Supplementary Material}
\input{sections/supplementary}

\end{document}

%% file: sections/01_introduction.tex
\def\thefootnote{*}\footnotetext{Both authors contributed equally to this research.}

Self-supervised pre-training models have achieved remarkable success in boosting the performance of Transformer-based architectures in Computer Vision. Inspired by the BERT model~\citep{devlin2019bert}, the Masked Image Modeling (MIM) pre-training objective~\citep{bao2022beit} has become widely adopted in literature as a powerful self-supervision method for vision tasks. In particular, this pre-training objective involves masking random image patches and reconstructing the corrupted visual input. While several recent studies have refined the MIM approach~\citep{chen2022context,he2022masked}, there has been little exploration of alternative pre-training objectives in the visual domain. In contrast, in the field of Natural Language Processing (NLP), several methods~\citep{song2020mpnet,yang2019xlnet} have surpassed the BERT pre-training objective with the introduction of different paradigms that aim to solve the drawbacks of previous methods.

Drawing inspiration from NLP, we investigate a permutation-based pre-training strategy, which we term Permuted Image Modeling (PIM). This approach autoregressively predicts permuted image patches maintaining contextual bi-directionality without corrupting any part of the input. Despite offering an improvement over the standard MIM-based objective, the autoregressive PIM technique reduces the amount of positional information available for each prediction. To tackle this issue, we propose a Masked and Permuted pre-training solution for Vision Transformers (\ours) which leverages auxiliary position information as input during pre-training, thus allowing the model to access the positional information of each image patch. 

In addition to the pre-training objective, a crucial aspect of self-supervised vision pre-training is the design of visual targets, used as supervisory signal. While some works have employed low-level and hand-crafted visual features~\citep{huang2022contrastive,liu2022swin,wei2022masked}, the dominant approach is to employ discrete visual tokens to reconstruct the corrupted input~\citep{bao2022beit,chen2022context,el2021large,peng2022beit}. In this context, although BEiT~\citep{bao2022beit} initially employed DALL-E~\citep{ramesh2021zero} visual tokens, its performance has been surpassed by VQ-KD, proposed in BEiT v2~\citep{peng2022beit}. In particular, VQ-KD employs an encoder-decoder architecture that reconstructs CLIP features and is directly trained on ImageNet-1k~\citep{deng2009imagenet}, requiring retraining to achieve satisfactory results on other datasets. In contrast to previous works, we propose $k$-CLIP, a novel discrete tokenizer for generating visual tokens that can directly employ CLIP features without requiring training an ad hoc discrete autoencoder.

\begin{figure*}[t]
    \centering
    \includegraphics[width=\textwidth]{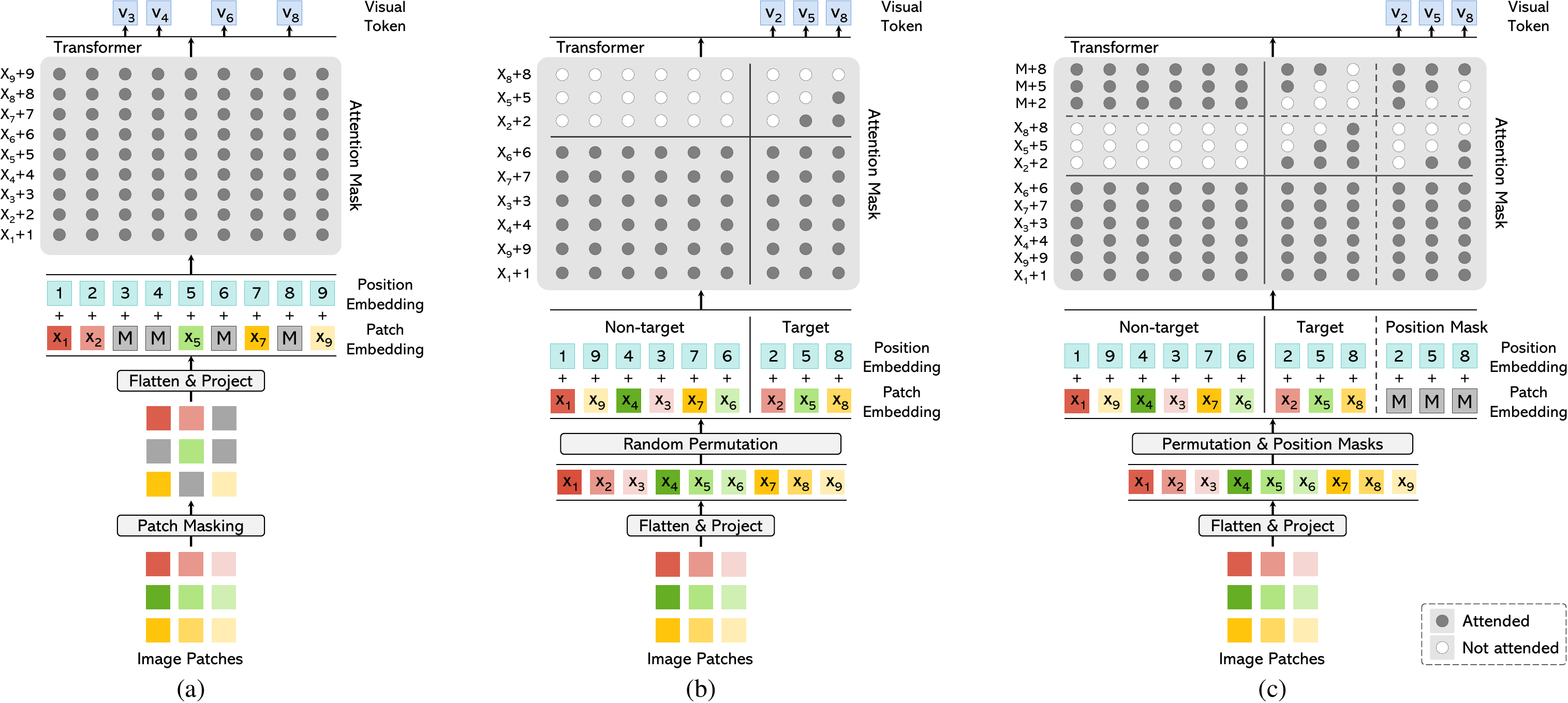}
    \vspace{-0.4cm}
    \caption{(a) Masked Image Modeling (MIM). (b) Permuted Image Modeling (PIM). (c) Masked and Permuted pre-training for Vision Transformers (\ours). While MIM reconstructs visual tokens from randomly masked image patches, PIM autoregressively predicts tokens associated with permuted image patches. \ours uses PIM to capture intra-patch dependency and takes auxiliary position information as input to ensure that the model sees a full sequence of patches at each target position.}
    \label{fig:mim_pim_mpit}
\end{figure*}

To evaluate the effectiveness of the proposed pre-training model and tokenizer, we conduct experiments that provide a fair comparison between models, adhering to the same experimental settings. This approach allows to accurately measure the efficacy of each model under consistent conditions and effectively compare their relative strengths and weaknesses. Experimental results demonstrate that our \ours model achieves competitive performance and surpasses both mask- and permutation-based image pre-training. Additionally, we show that the visual tokens extracted by our proposed $k$-CLIP tokenizer exhibit richer semantic information than competitors, outperforming both DALL-E and VQ-KD visual tokens when employed directly for image classification.

%% file: sections/02_related.tex
\tinytit{Self-supervised learning} 
Several solutions have been introduced in the last years to effectively pre-train vision-based architectures via self-supervised learning, initially based on different pretext tasks~\citep{doersch2015unsupervised,gidaris2018unsupervised,noroozi2016unsupervised,wang2015unsupervised,zhang2016colorful} and then exploiting contrastive learning paradigms~\citep{chen2020simple,chen2021exploring,grill2020bootstrap,he2020momentum,oord2018representation}. The advent of Vision Transformer models (ViT)~\citep{dosovitskiy2021image} has pushed towards the introduction of new increasingly sophisticated self-supervised pre-training strategies~\citep{bao2022beit,caron2021emerging,chen2020generative,chen2021empirical}. In this context, motivated by the great success in NLP, some attempts have been made to effectively adapt the Masked Language Modeling paradigm~\citep{devlin2019bert,liu2019roberta} and its auto-regressive variant~\citep{brown2020language} in the Computer Vision domain, either by directly predicting pixels~\citep{chen2020generative}, image patches~\citep{dosovitskiy2021image}, or discrete visual tokens~\citep{bao2022beit}.
In particular, the recently proposed BEiT approach~\citep{bao2022beit} effectively performs pre-training via image patch masking and predicts the discretized labels~\citep{ramesh2021zero} of masked patches. \rev{Since the introduction of BEiT, many subsequent methods based on similar pre-training strategies have been presented~\citep{chen2022context,dong2023peco,el2021large,he2022masked,tian2022beyond,tian2023integrally,wei2022masked,wei2022mvp,xie2022simmim,zhang2023hivit}. Some of these methods~\citep{chen2022context,el2021large,he2022masked} have introduced an encoder-decoder architecture to separate uncorrupted encoded information from masked tokens, which are employed directly as input to the decoder. 
\citet{zhang2023hivit} introduced HiViT, a simplified hierarchical Vision Transformer that combines the advantages of ViT and Swin~\citep{liu2022swin} models while being more efficient for Masked Image Modeling.~\citep{tian2023integrally} proposed integrally pre-trained Transformer pyramid networks, which jointly optimize a HiViT-based network backbone and a feature pyramid neck to minimize the transfer gap between pre-training and downstream tasks. Additionally,~\citep{tian2022beyond} investigated alternative learning objectives for token-based pre-training, providing insights into effective design principles beyond traditional masking approaches.} Inspired by findings from NLP literature~\citep{song2020mpnet,yang2019xlnet}, our proposed method aims to overcome the limitations of Masked Image Modeling in self-supervised pre-training.

\tit{Visual targets} In the domain of self-supervised learning, visual targets are a set of specific objectives that are used as the supervisory signal to pre-train visual models. Based on the type of signal, the targets can be categorized into low-level visual features, hand-crafted features, and visual tokens.
Some recent studies~\citep{fang2022corrupted,he2022masked,huang2022contrastive,liu2022swin,xie2022simmim} have utilized pixel information as the low-level supervisory signal for self-supervised pre-training. In contrast,~\citep{wei2022masked} have used HOG hand-crafted features to reconstruct the masked visual input. Recently, visual targets based on CLIP~\citep{radford2021learning} have been used with remarkable success, either by directly employing CLIP features~\citep{MILAN2022,wei2022mvp,zhang2022cae} or training a discrete tokenizer~\citep{peng2022beit} to reconstruct the semantic features encoded by CLIP. In our method, we propose a discrete visual tokenizer based on CLIP features which offers a novel approach to visual pre-training without the need for specific training over a particular dataset.

%% file: sections/03_preliminaries.tex
In this section, we detail two pre-training strategies that are the starting point of our proposal, \ie, Masked Image Modeling (MIM) and Permuted Image Modeling (PIM), and we introduce the terminology used in the rest of the paper.

\tit{Image patches}
In this study, we adopt ViT~\citep{dosovitskiy2021image} as the backbone network for our architecture. ViT splits an image into a sequence of 2D image patches, which are linearly projected to the model embedding space and elaborated through multiple attention blocks. Given an input image $\bm{x} \in \mathbb{R}^{H\times W\times C}$, this is mapped into a sequence of $N$ square patches $\{ \bm{x}_i^p \}_{i=1}^N$, where $\bm{x}_i^p\in \mathbb{R}^{P\times P \times C}$ is the $i$-th patch of the input image. Subsequently, a linear layer is applied to each flattened patch to project it to the input dimensionality of the model $D$, outputting the patch embeddings $\{ \bm{x}_i \}_{i=1}^N$, where $\bm{x}_i\in \mathbb{R}^{D}$, which are then added to the learnable 1D positional embeddings $\bm{E}_{\text{pos}} \in \mathbb{R}^{N \times D}$.

The ViT~\citep{dosovitskiy2021image} encoder consists of $L$ identical layers of Transformer blocks, where the output embeddings of the last layer represent the encoded representations of the $N$ input image patches. In our experiments, we consider an image dimension of $ 224 \times 224 $ with a patch dimension $P$ of $16 \times 16$, constituting an input sequence of $14 \times 14 = 196$ patch embeddings.

\tit{Visual tokens} In self-supervised pre-training, learned supervisory signals are typically used to effectively pre-train the visual backbone. In this work, we employ visual tokens as supervisory signals during the pre-training phase. Specifically, we represent targets as a discrete token sequence, accomplished by utilizing a visual tokenizer $\mathcal{T}(\bm{x})$ on the input image $\bm{x}$. The visual tokenizer maps the image pixels onto a visual codebook (or vocabulary), generating a sequence of tokens $\bm{v} = [v_1, \dots, v_N] \in \mathcal{V}^{(H/P) \times (W/P)}$, where $\mathcal{V}$ represents the vocabulary containing discrete token indices. Our approach employs a $14 \times 14$ grid of visual tokens to represent each image, while the vocabulary size is set to $|\mathcal{V}|=8192$.

\tit{Masked Image Modeling (MIM)} Inspired by the Masked Language Modeling strategy utilized in BERT~\citep{devlin2019bert}, the MIM paradigm is a pre-training technique for vision tasks that aims to recover visual information from a corrupted input image. This is achieved by randomly masking a portion of the image patches and predicting the visual tokens related to the corrupted region of the input, as depicted in Fig.~\ref{fig:mim_pim_mpit} (a). Like Masked Language Modeling, MIM has some inherent disadvantages. Firstly, a mask token $M$ is introduced during pre-training and never used during fine-tuning, leading to a discrepancy in the pre-training and fine-tuning phases. Secondly, given the masked tokens $\bm{\bar{x}}$ and the uncorrupted context $\bm{\tilde{x}}$, the probability $p(\bm{\bar{x}}\mid\bm{\tilde{x}})$ is typically factorized, assuming the independence of reconstructed patches. To address these issues, the NLP literature has investigated a permutation-based variant~\citep{yang2019xlnet} that can reduce the disadvantages of standard Masked Language Modeling. These results motivate us to investigate the application of this strategy to vision tasks.

\tit{Permuted Image Modeling (PIM)} The PIM self-supervised pre-training objective differs from MIM in two key components: the use of patch permutations and attention masking to capture bidirectional contexts. Specifically, PIM permutes patch embeddings, splits them into non-target and target patches, and predicts the visual tokens associated with the target patches using an auto-regressive approach. Attention masking is then applied to reduce the visibility of patches in the attention process, allowing a target patch to not access its contextual information (\ie, its content) during prediction while remaining visible to patches that come after it in the permuted order. A visual representation of the process is shown in Fig.~\ref{fig:mim_pim_mpit} (b). 

Formally, given an input image $\bm{x}$, we extract the patch embeddings $\{  \bm{x}_i \}_{i=1}^N$, and tokenize it into $N$ visual tokens $\{  v_i \}_{i=1}^N$. We define $\mathcal{Z}_t$ as the set of all $N!$ possible permutations of the length-$N$ index sequence $\{1,2,\dots, N\}$. 
Given a permutation $\bm{z}$, we use $z_t$ and $\bm{z}_{<t}$ to denote the $t$-th element and the first $t-1$ elements of $\bm{z}$, respectively.
After applying $\bm{z}$, the permuted patch embeddings are fed into an $L$-layer ViT backbone to extract the final hidden representations. For each input embedding $\bm{x}_{z_t}$ at position $z_t$ in the considered permutation $\bm{z}$, we use a softmax classifier to predict the corresponding visual token. The goal of PIM is to maximize the following log-likelihood objective:
\begin{equation}
\label{eq:pim}
    \max_{\theta} \sum_{\bm{x} \in \mathcal{D} }\mathbb{E}_{\bm{z} \in \mathcal{Z}_t}  \left [ \; \sum_{t=c+1}^{N} \log p_\theta(v_{z_t} | \bm{x}_{z_{<t}} ) \; \right ],
\end{equation}
where $\theta$ represents the model parameters, $\mathcal{D}$ denotes the training dataset, $\bm{x}_{z_{<t}}$ are the only patch embeddings that are visible at position $z_t$, and $c$ is a cutting point applied to split the permutation $\bm{z}$ in a subset of non-target patch embeddings $\bm{x}_{z_{\leq  c}}$ and target patch embeddings $\bm{x}_{z_{> c}}$. The aim of $c$ is to reduce the number of visual tokens to be predicted, thus mitigating optimization difficulties. 

While MIM preserves full positional information of each image patch, during the prediction process, PIM can access the contextual and positional information only of the $t-1$ patches preceding $z_t$, as shown in Eq.~\ref{eq:pim}. Given this lack of full positional information, PIM introduces an input discrepancy between pre-training and fine-tuning, underscoring the need for a pre-training technique that combines the advantages of both MIM and PIM while mitigating their respective limitations.

%% file: sections/04_method.tex
\begin{figure*}[t]
    \centering
    \includegraphics[width=\textwidth]{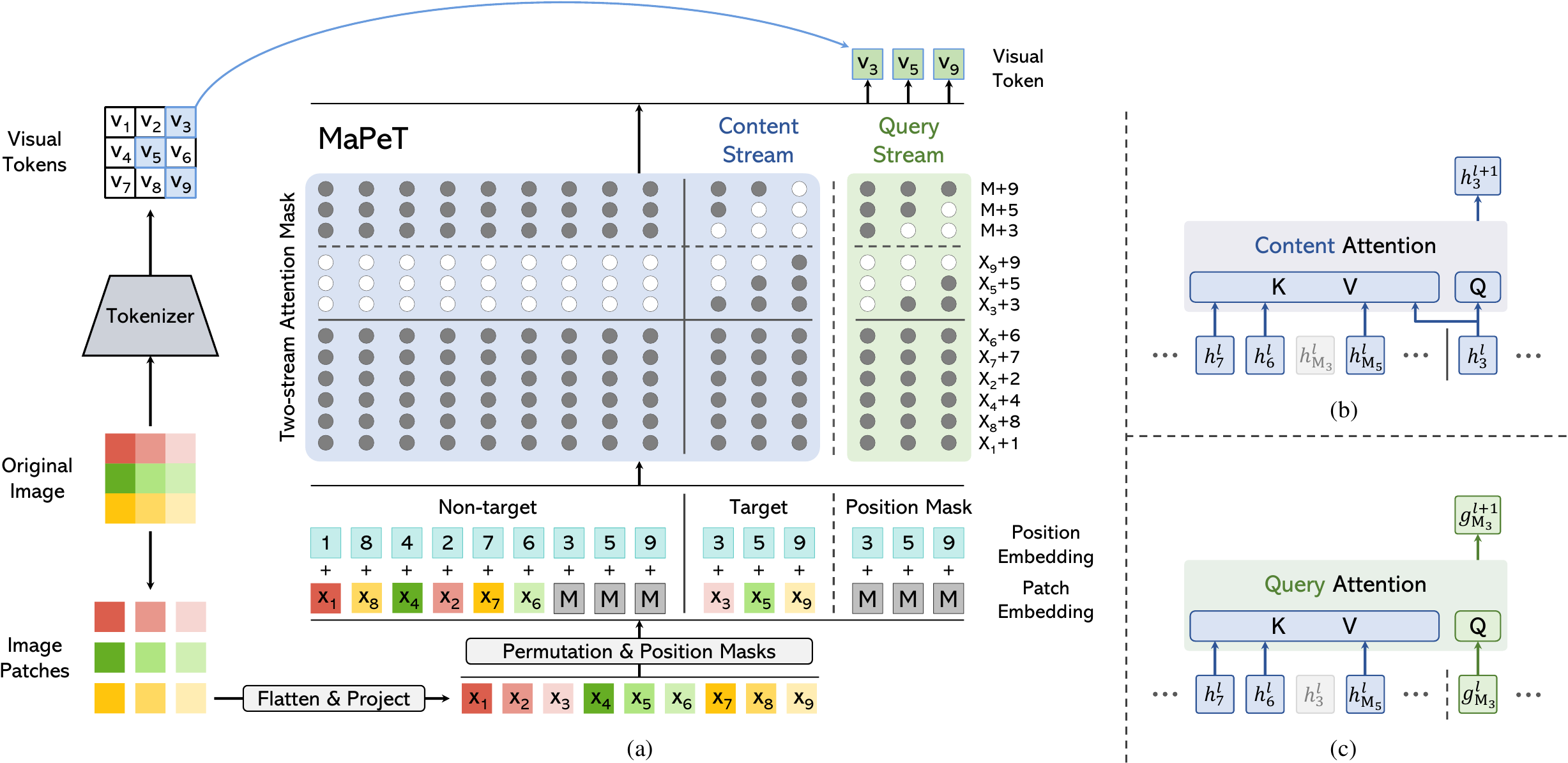}
    \vspace{-0.4cm}
    \caption{Overview of our \ours pre-training (a), with content stream attention (b), which follows the standard self-attention mechanism, and query stream attention (c), which lacks information about the content of the patch embedding $\bm{x}_{z_t}$ whose visual token $v_{z_t}$ is to be predicted. The blue and green masks in (a) are the content and query attention masks employed in the two-stream self-attention.
    }
    \label{fig:mpit}
\end{figure*}

In this section, we present \ours, a novel pre-training paradigm that combines masked and permuted image modeling strategies. In addition, we also introduce the $k$-CLIP visual tokenizer, which exploits discretized CLIP features to produce visual tokens.

\subsection{Masked and Permuted Pre-Training} 
Drawing inspiration from NLP literature~\citep{song2020mpnet,yang2019xlnet}, we propose a novel pre-training methodology called \textbf{Ma}sked and \textbf{Pe}rmuted Vision \textbf{T}ransformer (\ours) which builds on the strengths of both MIM and PIM to enhance performance on vision tasks. In particular, our approach overcomes the independence assumption of reconstructed patches, thus capturing intra-patch dependencies more effectively. Moreover, \ours incorporates auxiliary position embeddings during pre-training, enabling the model to access position information for all patches, thereby resolving the pre-training fine-tuning discrepancy introduced by PIM. Fig.~\ref{fig:mim_pim_mpit} (c) shows an overview of the \ours approach. 

Given a permutation $\bm{z}$ and a cutting point $c$, \ours can predict the visual token associated with an input patch embedding $\bm{x}_{z_t}$ by leveraging the content and position of the preceding $\bm{x}_{z_{<t}}$, as well as the position of the subsequent target embeddings $\bm{x}_{z_{>t}}$. 
To this end, we introduce the concept of learnable masked token $M \in \mathbb{R}^D$, which is used to express the positional information of $\bm{x}_{z_{>c}}$.
By repeating $N-c$ times the token $M$, we obtain $\{M_i\}_{i=c+1}^{N}$ identical masked tokens, which are then summed to the positional embedding $\{E_{\text{pos}}^i\}_{i=z_{c+1}}^{z_N}$ of each target $\bm{x}_{z_{> c}}$.
These resulting position-aware masked tokens $\bm{M}_{\text{pos}}=\{M_{c+1}+E_{\text{pos}}^{z_{c+1}}, \dots, M_{N}+E_{\text{pos}}^{z_N}\}$ are concatenated to the input patch embeddings $\bm{H}^0$ thus obtaining the augmented input  $\bm{H}^0_M = [\bm{H}^0, \bm{M}_{\text{pos}}]$. Note that the positional embeddings are permuted according to the same permutation $\bm{z}$ applied to the patches. For convenience, we introduce ${\bm{M}_{z_{\geq t}}}$ which represents the subset $\{M_{\text{pos}}^i\}_{i=\text{max}(1, \; t-c)}^{N-c}$. Intuitively, when $t \le c$, we have that ${\bm{M}_{z_{\geq t}}}$ comprises all $\bm{M}_{\text{pos}}$. Conversely, when $t>c$, ${\bm{M}_{z_{\geq t}}}$ only includes the position-aware masked tokens related to $\bm{x}_{z_{\geq t}}$. 

The training objective of our \ours model is to maximize the log-likelihood of predicting the visual token $v_{z_{t}}$ associated with the patch embedding $\bm{x}_{z_{t}}$ given $\bm{x}_{z_{<t}}$:
\begin{equation}
\label{eq:mpit}
    \max_\theta \sum_{\bm{x} \in \mathcal{D} }\mathbb{E}_{\bm{z} \in \mathcal{Z}_t} \left [ \, \sum_{t=c+1}^{N} \log p_\theta(v_{z_{t}} | \bm{x}_{z_{<t}}, \bm{M}_{z_{\geq t}}) \, \right].
\end{equation}

By doing so, \ours allows the patch embedding $\bm{x}_{z_{t}}$ to attend to contextual information of the patch embeddings ${\bm{x}_{z_{<t}}}$ as well as the positional information of ${\bm{x}_{z_{>t}}}$. This approach compensates for the position discrepancy of PIM and provides the model with information about the position of the target patches whose visual tokens are to be predicted. Fig.~\ref{fig:mpit} shows an illustration of our \ours method.


\tit{Two-stream self-attention pre-training} 
Since the target patch embeddings $\bm{x}_{z>c}$ follow the permuted order, the next predicted patch can occur in any position. As a consequence, masking the attention matrix of a ViT encoder, instead of corrupting the input like in MIM, makes the backbone architecture non-trivial. 
To implement PIM on a ViT backbone, we adopt the two-stream self-attention mechanism introduced by XLNet~\citep{yang2019xlnet}. Specifically, our model consists of two attention streams: a query stream and a content stream. 
The query stream $g_\theta(\bm{x}_{z_{<t}}, {\bm{M}_{z_{\geq t}}})$ accesses the content of the previous patches in the permuted sequence at a given position $t$. However, it does not access the content of $\bm{x}_{z_{t}}$, only viewing the position ${\bm{M}_{z_{\geq t}}}$ of the subsequent patches. In contrast, the content stream $h_\theta(\bm{x}_{z_{\leq t}}, {\bm{M}_{z_{> t}}})$ encodes the content of both the previous elements in the sequence and the element in position $z_t$, with positional information ${\bm{M}_{z_{> t}}}$ of the remaining patches. The input of the first query stream layer $\bm{g}^{(0)}$ consists of the masked elements $\bm{M}_{\text{pos}}$, which encode the position of the target patch embeddings $\bm{x}_{z_{> c}}$. Instead, the input of the first content stream layer $\bm{h}^{(0)}$ is the augmented input sequence $\bm{H}^0_M$.
Formally, for each Transformer layer $l$ with $l = 1, \ldots, L$, the attention mechanism of both content and query streams can be defined as:
\begin{equation}
\begin{aligned}
\label{eq:contentAttention}
   h^{(l)}_{z_{t}} \xleftarrow{} \mathsf{Attention}(\bm{Q}=h^{(l-1)}_{z_t}, \bm{KV}=(\bm{h}^{(l-1)}_{{z_{\leq t}}},\bm{h}^{(l-1)}_{{\bm{M}_{z_{>t}}}}) ; \, \theta), \\
    g^{(l)}_{z_{t}} \xleftarrow{} \mathsf{Attention}(\bm{Q}=g^{(l-1)}_{z_t}, \bm{KV}=(\bm{h}^{(l-1)}_{{z_{<t}}},\bm{h}^{(l-1)}_{{\bm{M}_{z_{\geq t}}}}) ; \, \theta),
\end{aligned}
\end{equation}
where $\bm{Q}$, $\bm{K}$, and $\bm{V}$ are respectively queries, keys, and values of the attention operator. Both query and content streams use separate attention masks to limit the visible contextual and positional information for each patch. During pre-training, the output of the query stream is used as the model output. At the fine-tuning step, the query stream is dropped, and only the content stream is used, returning to the standard ViT backbone.
Fig.~\ref{fig:mpit} shows the content stream (b), the query stream (c), and how they are integrated into our \ours architecture.

\tit{Attention masking} To limit the number of visible patches in the content and query attention operations, as described in Eq.~\ref{eq:contentAttention}, \ours leverages attention masking. In particular, the permutation $z$ influences the creation of two distinct attention masks: one for the content attention stream and one for the query attention stream. 
The content mask guarantees that only patch embeddings $\bm{x}_{z_{\leq t}}$ and positional tokens $\bm{M}_{z_{> t}}$ are visible to the patch embedding $\bm{x}_{z_t}$ in the content stream.
On the other hand, the query mask ensures that only patch embeddings $\bm{x}_{z_{<t}}$ and positional tokens $\bm{M}_{z_{\geq t}}$ are visible to $\bm{x}_{z_t}$ in the query stream. 

\subsection{$k$-CLIP: Discretized CLIP-based Tokenizer} 
\label{sec:kclip}
The role of visual tokenizers in pre-training pipelines is significant, as they provide crucial guidance for downstream fine-tuning outcomes. To possibly reduce the overhead of tokenizer retraining of previously proposed approaches~\citep{peng2022beit}, we explore the impact of directly utilizing discretized CLIP features. In particular, we propose a novel visual tokenizer, called $k$-CLIP, that employs discretized CLIP features as visual tokens.  Our method does not rely on any pre-training or supervised data to create the visual tokens, thus enabling pre-training without access to large amounts of labeled data or a particular pre-training dataset. Furthermore, the use of CLIP features enables our method to capture high-level visual semantics that are more meaningful for downstream tasks, further improving the performance of the learned representations.

Specifically, we sample visual features from the ImageNet dataset~\citep{deng2009imagenet} using the CLIP model~\citep{radford2021learning} and cluster them using $k$-means to obtain $|\mathcal{V}|=8192$ centroids. During pre-training, these centroids are indexed to retrieve the corresponding visual tokens for prediction.
Formally, our visual tokenizer $\mathcal{T}(\bm{x})$ consists of a ResNet-based CLIP visual encoder and a $k$-means model. 
The visual encoder $f_v: \mathbb{R}^{H \times W \times C} \rightarrow \mathbb{R}^{H_c \times W_c \times D_c}$ maps an input image $\bm{x}$ to a grid of visual features that correspond to the activations produced by the last convolutional layer of the CLIP backbone. 
The visual features are then reshaped to a $N_c \times D_c$ matrix, where $N_c = H_c \times W_c$. 
In our method, the CLIP visual encoder generates visual features of shape ${N_c \times D_c} = {196 \times 4096}$, which are subsequently indexed by $k$-means and mapped to a sequence of $196$ discrete visual tokens $\bm{v} = [v_1, \dots, v_N] \in \mathcal{I}$.
The set of visual tokens $\bm{v} = \{v_i\}_{i=1}^N$ is defined by the $k$-means centroid indexes $\mathcal{I}=\{1, \dots, |\mathcal{V}|\}$.
To mitigate the computational complexity, we randomly sample $2\%$ of the approximately $250$ million $D_c$-dimensional CLIP features extracted from ImageNet. The sampled feature collection is then used to fit the $k$-means clustering model.

%% file: sections/05_experiments.tex
\subsection{Experimental Setup}
Self-supervised pre-training literature has introduced several training and fine-tuning procedures that encompass different tokenizers, hyperparameters, and visual backbones. To ensure a fair and unbiased comparison among these pre-training algorithms, we opt to evaluate our \ours method, as well as the related approaches in the literature, using identical experimental configurations. This strategy allows for the isolation of the algorithmic factor in the experiments and promotes unbiased comparisons of pre-training objectives. To assess the effectiveness of the pre-trained objective under consideration, we fine-tune our models on a downstream classification task. \rev{Additionally, following recent literature, we perform fine-tuning experiments for semantic segmentation, to assess the generalization of the proposed pre-training objective to dense predictions.}

\input{tables/table_nlp}
\input{tables/table_comparison}

\tit{Pre-training setup} 
We first investigate the influence of the visual tokenizer and evaluate the proposed $k$-CLIP against VQ-KD~\citep{peng2022beit}. To minimize the computational effort, we pre-process the ImageNet-1k training dataset for both the tokenizers and store the visual tokens associated with each image. For this reason, our pre-training augmentation policy only includes color jittering to preserve image patch positions corresponding to the pre-extracted visual tokens.

We compare \ours against different pre-training objectives. In particular, we employ a ViT-based backbone that is pre-trained according to the pure MIM objective formulation. Note that this differs from the pre-training strategy proposed by the BEiT approach~\citep{bao2022beit} as it lacks the block-wise masking algorithm, which progressively extracts multiple blocks of patches until 40\% of the positions are masked. In our MIM-based pre-training, we replace the block-wise masking strategy with a random patch masking approach, thus keeping it similar to a BERT-like solution~\citep{devlin2019bert} applied to Computer Vision tasks. Analogously, we also pre-train a ViT backbone through a double-stream architecture according to the PIM objective described in Sec.~\ref{sec:preliminaries}. Moreover, we consider the standard BEiT model~\citep{bao2022beit} (\ie, a MIM-based pre-training with block-wise masking) and CAE~\citep{chen2022context}. The CAE method employs an encoder-decoder architecture where the encoder processes only visible image patches (50\% of the entire image), while the remaining 50\% is masked. A latent contextual regressor predicts the masked representation based on the encoder output, and a lightweight decoder processes the output of the regressor, which is then used to predict the visual token of the related masked patches.  

\tit{Image classification setup} During the classification fine-tuning stage, the final hidden layer of the ViT-based backbone extracts features that are then combined via average pooling to generate a global image representation. This representation is subsequently fed into a softmax classifier. Following the linear probing experiment reported in~\citep{bao2022beit}, we also train a linear classifier head over the output representation produced by the frozen pre-trained backbone.

We design three different model variants based on Vision Transformer~\citep{dosovitskiy2021image}, \ie, ViT-Tiny (ViT-T), ViT-Small (ViT-S), and ViT-Base (ViT-B). Our \ours model is trained by setting the cutting point $c$ to $50$, $50$, and $60$, respectively. We refer the reader to the supplementary material for detailed pre-training and fine-tuning hyperparameters.

\tit{\rev{Semantic segmentation setup}} 
\rev{Semantic segmentation is a pixel-wise classification task that predicts semantic labels for each input image pixel. Our experimental framework follows the setting proposed in BEiT v2~\citep{peng2022beit} and utilizes the ADE20K dataset~\citep{zhou2017scene,zhou2019semantic} as a comprehensive benchmark comprising 25k images spanning 150 semantic categories. For the segmentation architecture, we employ UperNet~\citep{xiao2018unified} task layer and fine-tune the model for 160k iterations with input images resized to a resolution of $512 \times 512$. The results are reported in terms of mean Intersection over Union (mIoU). To facilitate reproducibility, we include a detailed list of hyperparameters in the supplementary material.}

\tit{Computational requirements} The pre-training experiments conducted with our \ours model involve the utilization of different GPU configurations. Specifically, the ViT-T, ViT-S, and ViT-B models require the deployment of 16, 32, and 64 GPUs, respectively. The pre-training process for each model takes approximately one day to complete. In contrast, during the fine-tuning phase, we employ 4 GPUs for the ViT-T and ViT-S models, and 16 GPUs for the ViT-B model. The fine-tuning duration for each model is 48 hours, 36 hours, and 12 hours, respectively. For all experiments, we utilize an NVIDIA V100 GPU architecture with 16GB of memory.

\subsection{Experimental Results}
\tit{Pre-training objectives comparison} To validate the assumptions made on the pre-training objectives presented in Section~\ref{sec:method}, we conduct a comparison of MIM, PIM, and our proposed \ours, as shown in Table~\ref{tab:comparison_nlp}. This comparison evaluates the top-1 accuracy and linear probe accuracy of these approaches.
The results indicate that MIM pre-training is less effective compared to PIM due to the input noise introduced by masked tokens and the independent reconstruction of patches. PIM outperforms MIM in most comparisons. For instance, when employing ViT-T with $k$-CLIP and ViT-S with VQ-KD, PIM achieves improvements of $0.4\%$ and $0.1\%$ in classification accuracy, respectively, over MIM. However, PIM demonstrates comparable performance to MIM when applied to the ViT-B backbone.
Furthermore, our proposed \ours consistently outperforms PIM in nearly all cases. It exhibits accuracy gains of $0.2\%$, $0.3\%$, and $0.7\%$ in top-1 accuracy, as well as $2.8\%$, $2.1\%$, and $0.5\%$ in linear probe accuracy respectively on ViT-T and ViT-S when using the $k$-CLIP tokenizer, and on ViT-B when employing the VQ-KD tokenizer. These findings underscore the significance of addressing position inconsistency between pre-training and fine-tuning, particularly in the context of permutation-based image pre-training.

\tit{Comparison with state-of-the-art models} Table~\ref{tab:comparison} presents a comprehensive analysis of the performance of \ours, BEiT, and CAE in terms of top-1 accuracy and linear probe accuracy across all ViT-based backbones.
Firstly, our pre-trained \ours model showcases significant performance improvements across all Tiny, Small, and Base backbones compared to ViT-based models trained with random initialization, as evidenced in Table~\ref{tab:comparison}.
Secondly, our results demonstrate that both BEiT and \ours outperform CAE. We hypothesize that the CAE encoder, which only observes $50\%$ of the total sequence during pre-training, may suffer from position discrepancies between the pre-training and fine-tuning phases.
Thirdly, it is noteworthy that BEiT improves overall results compared to MIM in Table~\ref{tab:comparison_nlp}. We attribute this improvement to the blockwise masking technique employed by BEiT. This technique follows the principle of image spatial locality, which posits that adjacent patches exhibit similarities in terms of visual information. By employing blockwise masking, the density of uncorrupted visual content is increased, while the noise introduced by masked tokens is concentrated in fewer locations instead of being sparsely distributed.
Furthermore, \ours consistently outperforms all competitors across the three model variants when employing the $k$-CLIP tokenizer. Specifically, our model achieves top-1 accuracy margins of $0.7$\%, $0.2$\%, and $0.3$\% against BEiT on ViT-T, ViT-S, and ViT-B, respectively, while exhibiting margins of $1.8$\%, $0.6$\%, and $0.9$\% against CAE across the same three backbones. In contrast, \ours demonstrates comparable results to BEiT when employing the VQ-KD tokenizer on the ViT-S and ViT-B backbones.

\input{tables/table_sota}

We also report the comparison of our best variants and other state-of-the-art self-supervised pre-training models in Table~\ref{tab:sota}. Note that the variability in the experimental settings and the different supervisory signals used in each approach may affect the fairness of the comparisons.
Notably, although current literature has not extensively explored performance benchmarking on the ViT-S backbone, the Small version of our \ours model outperforms BEiT~\citep{bao2022beit} and CAE~\citep{chen2022context} by $0.5\%$ and $0.2\%$ respectively, yielding the best performance for both $k$-CLIP and VQ-KD. In the case of the \mbox{ViT-B} backbone, \ours surpasses most of the considered approaches especially when using the VQ-KD tokenizer, except for BEiT v2~\citep{peng2022beit} which gets slightly better results. This performance gap can be explained by the lack of data augmentation in our model pre-training, which can significantly increase performance but at a higher computational cost.

The results in Table~\ref{tab:sota} are obtained after pre-training the models for 300 epochs. For completeness, in the lower part, we report the performance of other methods when pre-trained for a considerably large number of epochs (\ie, 800 and 1600). While a direct comparison using different pre-training epochs may not be completely informative, \ours still proves to perform better than other methods pre-trained for a larger number of epochs such as BEiT and CAE. 

\input{tables/table_segmentation}
\input{tables/suppl_table_new_domains}

\tit{\rev{Semantic segmentation results}} 
\rev{To evaluate the effectiveness of \ours on dense prediction tasks, we present its performance on the semantic segmentation benchmark ADE20K using the UperNet framework with the VQ-KD tokenizer. Table~\ref{tab:segmentation} presents the mIoU scores achieved with various ViT backbones and pre-training strategies. Notably, \ours surpasses both MIM and PIM objectives, delivering significant improvements of $+1.4\%$ and $+0.9\%$ on ViT-S, and $+1.1\%$ and $+1.4\%$ on ViT-B, respectively.
Compared to CAE~\citep{chen2022context} and BEiT~\citep{bao2022beit}, \ours achieves the highest mIoU across ViT-S and ViT-B backbones, while maintaining competitive performance on ViT-T. 
These results underscore the superiority of \ours over existing pre-training objectives and self-supervised approaches, particularly in addressing positional inconsistencies during pre-training and fine-tuning, which is critical for dense tasks like semantic segmentation. The robust generalization demonstrated by \ours highlights its potential as an effective pre-training strategy for segmentation tasks.}

\tit{Cross-domain transfer learning}
\label{sec:cross_domain} As an additional analysis, we examine the generalization capabilities of our proposed self-supervised pre-training technique, compared to BEiT~\citep{bao2022beit} and CAE~\citep{chen2022context}. All the considered architectures employ the VQ-KD visual tokenizer~\citep{peng2022beit} and undergo pre-training on the ImageNet-1k dataset~\citep{deng2009imagenet}. 
Subsequently, they are fine-tuned on three distinct data domains, namely Stanford-Cars~\citep{krause20133d}, Food-101~\citep{bossard2014food}, and FGVC-Aircraft~\citep{maji13fine-grained}. These datasets are chosen to represent diverse and real-world scenarios, ranging from object recognition in the automotive domain to food and aircraft classification. By employing a linear probe evaluation, we can quantitatively measure the ability of our pre-trained model to transfer knowledge and adapt to new tasks without fine-tuning.

The results, presented in Table~\ref{tab:suppl_new_domains}, clearly demonstrate the superior performance of \ours across all considered datasets. Our model outperforms both BEiT and CAE, highlighting its robustness and efficacy in capturing meaningful visual representations. 
These findings not only underline the potential of \ours as a powerful pre-training technique but also emphasize its cross-domain transfer learning capabilities, which enable practical relevance in various real-world applications.

\tit{Reconstruction ratio analysis}  
Here we discuss the relationship between the reconstruction ratio employed in \ours and its impact on image classification results. The reconstruction ratio indicates the proportion of target patches in relation to the entire input sequence. 
An advantage of \ours is that it attends to the target patch $\bm{x_{z_t}} \in \bm{x}_{z_{> c}}$ during the prediction of $\bm{x_{z_{t+1}}}$, allowing for the cutting point $c$ to be positioned at every value in the interval $c \in [1 \ldots N-1]$. 
This unique feature of \ours enables the potential reconstruction of all patches in an incremental and randomized manner, from the first to the last patch. On the other hand, models like BEiT and CAE lack the ability to reconstruct all image patches because this would lead to a masking ratio that is too high, resulting in the loss of visual context necessary for the reconstruction of the masked elements.

\input{tables/table_least_number}

Table~\ref{tab:ablation} refers to the analysis of the reconstruction ratio on the ViT-B architecture. It can be noted that \ours exhibits a preference for a reconstruction ratio of approximately 70\% of the complete sequence. Indeed, the initial target patches are compelled to establish correlations with a small portion of visible patches (namely, 30\% of the entire sequence) randomly distributed throughout the image. 
The model can acquire long-range spatial dependencies that are contingent on the position of the target patch in the permuted order. As the final patches attend to the majority of the visual content, they are likely to concentrate more on neighboring image patches, consistent with the principle of spatial locality. A reduction in the reconstruction ratio leads to an increase in the number of visible patches attended by the target patches that occur early in the permuted order. This results in an increase in the likelihood of having neighboring visible patches, which, in turn, diminishes the learned long-range spatial dependencies. In contrast, an excessively high reconstruction ratio of 85\% makes the pre-training objective of \ours excessively difficult, as early predictions are likely to be arbitrary, having access to only 15\% of the overall visual information. Consequently, excessively high or low reconstruction ratios impair the performance.

\begin{figure*}[t]
    \begin{center}
    \includegraphics[width=0.98\textwidth]{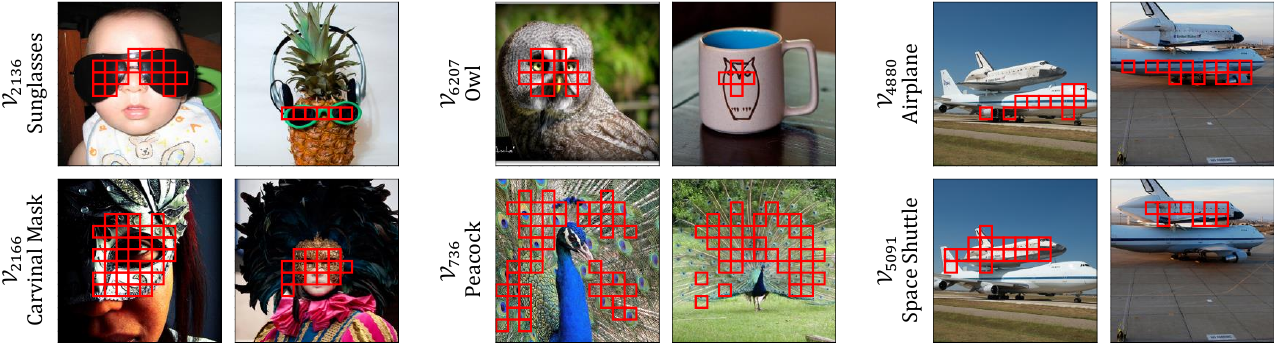}
    \end{center}
    \vspace{-.4cm}
    \caption{Visualization of image patches corresponding to the discrete tokens of our $k$-CLIP codebook, where semantically similar patches consistently share the same discrete token.}
    \label{fig:kclip_qualitatives}
    \vspace{-.1cm}
\end{figure*}

\subsection{Visual Tokenizer Analysis}
\label{sec:tokenizer}
In this section, we present a comprehensive qualitative and quantitative analysis of our proposed $k$-CLIP and compare its performance to the existing DALL-E~\citep{ramesh2021zero} and VQ-KD~\citep{peng2022beit} tokenizers.

\tit{Impact of tokenizer on model performance} Analyzing the results in Table~\ref{tab:comparison_nlp}, it can be noted that $k$-CLIP outperforms VQ-KD across all models evaluated on ViT-T, with improvements in top-1 accuracy of $1.0\%$, $0.4\%$, and $0.3\%$ observed on \ours, PIM, and MIM, respectively.
However, the results differ when evaluated on ViT-S and ViT-B, where VQ-KD exhibits better classification performance.
We argue that the simpler semantic features of $k$-CLIP, with their inherent correlation with the semantic density of the image, can serve as a more effective self-supervisory signal for smaller models. On the other hand, the VQ-KD codebook is trained specifically to reconstruct CLIP features, making it a more demanding pre-training signal and thus more suitable for larger models.

\tit{Image classification with discrete visual tokens} In this section, we analyze image classification performance when using discretized visual tokens directly as input to the model. We conduct this analysis using the previously employed tokenizers (\ie, our $k$-CLIP and VQ-KD~\citep{peng2022beit}) and also include the DALL-E visual tokenizer~\citep{ramesh2021zero}, which consists of a discrete variational autoencoder model. Specifically, we map the sequence of visual tokens to an embedding space of dimension $D$ through learnable embeddings trained in conjunction with two different model backbones: a lightweight MLP classification head and a ViT-Tiny model. The MLP head comprises an embedding layer, a linear layer with output dimension $D=192$, a ReLU activation, an average pooling operation, and an additional linear layer that projects the pooled features to the number of classes (\ie, 1000).

The results shown in Table~\ref{tab:tokenizers} indicate that the use of DALL-E visual tokens yields limited accuracy in this setting. This can be attributed to the relatively low amount of semantic information derived from the reconstruction task that DALL-E is trained on. In contrast, $k$-CLIP demonstrates significant performance superiority over alternative tokenizers when employed with both the MLP head and ViT-Tiny backbone. This observed trend can be attributed to the rich semantic information inherent in CLIP visual features, which is effectively preserved through the process of $k$-means discretization. Consequently, a strong correlation between visual tokens and image classes can be established, leading to favorable performance outcomes even when utilizing small models such as the MLP head.

\input{tables/table_tokenizers}

\tit{Codebook visualization}
Fig.~\ref{fig:kclip_qualitatives} presents a collection of examples that showcase the semantic associations between image patches and visual tokens, extracted using our \mbox{$k$-CLIP} tokenizer. 
This visualization effectively demonstrates the efficacy of our tokenizer in accurately capturing and representing the semantic content of the images within the considered dataset and shows its ability to recognize and group image patches that share common visual features and semantic meaning. As it can be noticed, the visual tokens shown are congruent with specific semantic concepts, albeit resulting in distinct representations for similar visual elements such as \texttt{sunglasses} and \texttt{carnival mask}, \texttt{owl} and \texttt{peacock}, or \texttt{airplane} and \texttt{space shuttle}.
Furthermore, it is worth noting that the visual tokens remain resilient to variations in color, style, rotation, and size. This is exemplified by the middle example, where the image depicts an owl print on a cup, yet it is still accurately identified using the \texttt{owl} visual token.
These observations emphasize the ability of our \mbox{$k$-CLIP} tokenizer to effectively identify and classify complex visual concepts.

%% file: tables/table_nlp.tex
\begin{table*}[t]
\centering
\setlength{\tabcolsep}{.45em}
\footnotesize
\resizebox{0.75\linewidth}{!}{
\begin{tabular}{lcc cc c cc c cc}
\toprule
& & & \multicolumn{2}{c}{\textit{ViT-T}} & & \multicolumn{2}{c}{\textit{ViT-S}} & & \multicolumn{2}{c}{\textit{ViT-B}} \\
\cmidrule{4-5} \cmidrule{7-8} \cmidrule{10-11}
& & & \textbf{Top-1} & \textbf{Linear Probe} & & \textbf{Top-1} & \textbf{Linear Probe} & & \textbf{Top-1} & \textbf{Linear Probe} \\
\textbf{Method} & \textbf{Tokenizer} & & \textbf{Acc. (\%)} & \textbf{Acc. (\%)} & & \textbf{Acc. (\%)} & \textbf{Acc. (\%)} & & \textbf{Acc. (\%)} & \textbf{Acc. (\%)} \\
\midrule
MIM & VQ-KD & & 74.6 & 55.3 & & 82.0 & \textbf{69.9} & & 83.8 & 72.3 \\
\textbf{PIM} & VQ-KD & & \textbf{74.9} & 56.4 & & 82.1 & 68.7 & & 83.7 & 73.3 \\
\rowcolor{LightCyan}
\textbf{\ours} & VQ-KD & & 74.5 & \textbf{59.6} & & \textbf{82.2} & 69.8 & & \textbf{84.4} & \textbf{73.8} \\
\midrule
MIM & $k$-CLIP & & 74.9 & 60.4 & & 82.0 & 71.1 & & 83.3 & 72.3 \\
\textbf{PIM} & $k$-CLIP & & 75.3 & 59.7 & & 81.8 & 69.5 & & 83.3 & \textbf{73.5} \\
\rowcolor{LightCyan}
\textbf{\ours} & $k$-CLIP & & \textbf{75.5} & \textbf{62.5} & & \textbf{82.1} & \textbf{71.6} & & \textbf{83.6} & \textbf{73.5} \\
\bottomrule
\end{tabular}
}
\caption{Fine-tuning results of different pre-training objectives in terms of top-1 and linear probe accuracy on ImageNet-1k. We report the results using both the VQ-KD and $k$-CLIP tokenizers.}
\label{tab:comparison_nlp}
\end{table*}

%% file: tables/table_comparison.tex
\begin{table*}[t]
\centering
\footnotesize
\setlength{\tabcolsep}{.3em}
\resizebox{0.8\linewidth}{!}{
\begin{tabular}{lcc cc c cc c cc}
\toprule
& & & \multicolumn{2}{c}{\textit{ViT-T}} & & \multicolumn{2}{c}{\textit{ViT-S}} & & \multicolumn{2}{c}{\textit{ViT-B}} \\
\cmidrule{4-5} \cmidrule{7-8} \cmidrule{10-11}
& & & \textbf{Top-1} & \textbf{Linear Probe} & & \textbf{Top-1} & \textbf{Linear Probe} & & \textbf{Top-1} & \textbf{Linear Probe} \\
\textbf{Method} & \textbf{Tokenizer} & & \textbf{Acc. (\%)} & \textbf{Acc. (\%)} & & \textbf{Acc. (\%)} & \textbf{Acc. (\%)} & & \textbf{Acc. (\%)} & \textbf{Acc. (\%)} \\
\midrule
ViT~\citep{dosovitskiy2021image} & - & & 73.7 & - & & 79.8 & - & & 81.8 & - \\
\midrule
CAE~\citep{chen2022context} & VQ-KD  & & 73.4 & 52.4 & & 81.6 & 63.4 & & 83.5 & 69.4  \\
BEiT~\citep{bao2022beit} & VQ-KD  & & \textbf{75.0} & \textbf{62.0} & & \textbf{82.2} & \textbf{72.6} & & \textbf{84.4} & \textbf{75.0} \\
\rowcolor{LightCyan}
\textbf{\ours} & VQ-KD & & 74.5 & 59.6 & & \textbf{82.2} & 69.8 & & \textbf{84.4} & 73.8 \\
\midrule
CAE~\citep{chen2022context} & $k$-CLIP & & 73.7 & 53.9 & & 81.5 & 62.9 & & 82.7 & 67.6 \\
BEiT~\citep{bao2022beit} & $k$-CLIP & & 74.8 & 61.5 & & 81.9 & 71.1 & & 83.3 & 73.3  \\
\rowcolor{LightCyan}
\textbf{\ours} & $k$-CLIP & & \textbf{75.5} & \textbf{62.5} & & \textbf{82.1} & \textbf{71.6} & & \textbf{83.6} & \textbf{73.5} \\
\bottomrule
\end{tabular}
}
\caption{Fine-tuning results compared to other self-supervised pre-training approaches, in terms of top-1 and linear probe accuracy.}
\label{tab:comparison}
\end{table*}

%% file: tables/table_sota.tex
\begin{table}[t]
\centering
\footnotesize
\setlength{\tabcolsep}{.25em}
\resizebox{\linewidth}{!}{
\begin{tabular}{lc cc ccc}
\toprule
& & & & \textit{ViT-S} & & \textit{ViT-B} \\
\cmidrule{5-5} \cmidrule{7-7}
\textbf{Method} &  & \textbf{\# Epochs} & & \textbf{Top-1 Acc. (\%)} & & \textbf{Top-1 Acc. (\%)} \\
\midrule
BEiT~\citep{bao2022beit} & & 300 & & 81.7 & & 82.9 \\
CAE~\citep{chen2022context} & & 300 & & 82.0 & & 83.6 \\
SplitMask~\citep{el2021large} & & 300 & & - & & 83.6 \\
MaskFeat~\citep{wei2022masked} & & 300 & & - & & 83.6 \\
PeCo~\citep{dong2023peco} & & 300 & & - & & 84.1 \\
MVP~\citep{wei2022mvp} & & 300 & & - & & 84.4 \\
BEiT v2~\citep{peng2022beit} & & 300 & & - & & \textbf{85.0} \\
\rowcolor{LightCyan}
\textbf{\ours ($\mathbf{k}$-CLIP)} & & 300 & & 82.1 & & 83.6 \\
\rowcolor{LightCyan}
\textbf{\ours (VQ-KD)} & & 300 & & \textbf{82.2} & & 84.4 \\
\midrule
\textcolor{gray}{BEiT~\citep{bao2022beit}} & & \textcolor{gray}{800} & & \textcolor{gray}{-} & & \textcolor{gray}{83.2} \\
\textcolor{gray}{CAE~\citep{chen2022context}} & & \textcolor{gray}{800} & & \textcolor{gray}{-} & & \textcolor{gray}{83.8} \\
\textcolor{gray}{CAE~\citep{chen2022context}} & & \textcolor{gray}{1600} & & \textcolor{gray}{-} & & \textcolor{gray}{83.9} \\
\textcolor{gray}{BEiT v2~\citep{peng2022beit}} & & \textcolor{gray}{1600} & & \textcolor{gray}{-} & & \textcolor{gray}{85.5} \\
\bottomrule
\end{tabular}
}
\caption{Comparison with state-of-the-art self-supervised pre-training models in terms of top-1 accuracy on ImageNet-1k. Results of competitors are extracted from the original papers.
}
\label{tab:sota}
\end{table}

%% file: tables/table_segmentation.tex
\begin{table}[t]
\centering
\footnotesize
\setlength{\tabcolsep}{.45em}
\resizebox{0.88\linewidth}{!}{
\begin{tabular}{>{\color{black}}l >{\color{black}}c c>{\color{black}}c c>{\color{black}}c c>{\color{black}}c}
\toprule
& & \multicolumn{6}{c}{\rev{\textbf{mIoU (\%)}}} \\
\cmidrule{4-8}
\textbf{Method} & \textbf{Tokenizer} & &  \textit{ViT-T} & & \textit{ViT-S} && \textit{ViT-B} \\
\midrule
MIM & VQ-KD & & 39.0 & & 45.6 & & 49.3 \\
PIM & VQ-KD  & & \textbf{40.2} & & 46.1 & & 49.0 \\
\rowcolor{LightCyan}
\textbf{\ours} & VQ-KD & & 39.3 & & \textbf{47.0} & & \textbf{50.4} \\
\midrule
CAE~\citep{chen2022context} & VQ-KD & & 39.2 & & 45.9 & & 50.2 \\
BEiT~\citep{bao2022beit} & VQ-KD  & & \textbf{39.9} & & 46.7 & & 50.3 \\
\rowcolor{LightCyan}
\textbf{\ours} & VQ-KD & & 39.3 & & \textbf{47.0} & & \textbf{50.4} \\
\bottomrule
\end{tabular}
}
\caption{\rev{Semantic segmentation results in terms of mIoU on ADE20K, comparing \ours with alternative pre-training objectives (top) and other self-supervised pre-training approaches (bottom).}}
\label{tab:segmentation}
\end{table}

%% file: tables/suppl_table_new_domains.tex
\begin{table}[t]
\centering
\footnotesize
\setlength{\tabcolsep}{.15em}
\resizebox{\linewidth}{!}{
\begin{tabular}{l c cc cc cc}
\toprule
& & \multicolumn{6}{c}{\textbf{Linear Probe Acc. (\%)}} \\
\cmidrule{4-8}
\textbf{Method} & \textbf{Tokenizer} & &  \textit{Stanford-Cars} & & \textit{Food-101} && \textit{FGVC-Aircraft} \\
\midrule
CAE~\citep{chen2022context} & VQ-KD  & & 53.6 & & 81.4  & & 40.5 \\
BEiT~\citep{bao2022beit} & VQ-KD  & & 64.5 & & \textbf{86.9} & & 44.9 \\
\rowcolor{LightCyan}
\textbf{\ours} & VQ-KD & & \textbf{68.5} & &  \textbf{86.9} & & \textbf{46.8}\\
\bottomrule
\end{tabular}
}
\caption{Comparison of model performance on cross-domain transfer learning. We measure the linear probe accuracy of our proposed \ours model compared to BEiT and CAE on three distinct data domains: Stanford-Cars~\citep{krause20133d}, Food-101~\citep{bossard2014food}, and FGVC-Aircraft~\citep{maji13fine-grained}.
}
\label{tab:suppl_new_domains}
\end{table}

%% file: tables/table_least_number.tex
\begin{table}[t]
\centering
\footnotesize
\setlength{\tabcolsep}{.3em}
\resizebox{0.95\linewidth}{!}{
\begin{tabular}{cc cc cc c}
\toprule
 & &  & & & & \textbf{Top-1} \\
\textbf{Cutting Point $c$} & & \textbf{Reconstruction Ratio} & & \textbf{Tokenizer} & & \textbf{Accuracy (\%)} \\
\midrule
30 && 85\% && VQ-KD && 84.2 \\
40 && 80\% && VQ-KD && 84.3 \\
50 && 75\% && VQ-KD && 84.3 \\
60 && 70\% && VQ-KD && \textbf{84.4} \\
98 && 50\% && VQ-KD && 84.1 \\
\bottomrule
\end{tabular}
}
\caption{Ablation study on the impact of the reconstruction factor on the top-1 fine-tuning accuracy of pre-trained ViT-B under the VQ-KD setting.}
\label{tab:ablation}
\end{table}

%% file: tables/table_tokenizers.tex
\begin{table}[t]
\centering
\footnotesize
\setlength{\tabcolsep}{.4em}
\resizebox{0.8\linewidth}{!}{
\begin{tabular}{lc cc c cc}
\toprule
& & \multicolumn{2}{c}{\textit{MLP}} & & \multicolumn{2}{c}{\textit{ViT-T}} \\
\cmidrule{3-4} \cmidrule{6-7}
 & & \textbf{Top-1} & \textbf{Top-5} & & \textbf{Top-1} & \textbf{Top-5} \\
\textbf{Tokenizer} & & \textbf{Acc. (\%)} & \textbf{Acc. (\%)} & & \textbf{Acc. (\%)} & \textbf{Acc. (\%)} \\
\midrule
DALL-E & & 4.1 & 11.7 & & 9.1 & 21.9 \\
VQ-KD & & 68.1 & 89.8 & & 72.3 & 92.2 \\
\rowcolor{LightCyan}
\textbf{$\mathbf{k}$-CLIP} & & \textbf{72.8} & \textbf{93.2} & & \textbf{76.2} & \textbf{94.9} \\
\bottomrule
\end{tabular}
}
\caption{Accuracy of image classification when employing visual tokens as model input. Note that using CLIP features for zero-shot image classification lead to a top-1 accuracy of 73.6\%.}
\label{tab:tokenizers}
\vspace{-.1cm}
\end{table}

%% file: sections/06_conclusion.tex
This paper presents \ours, a novel self-supervised pre-training approach designed for vision tasks. Our model effectively tackles the limitations of standard Masked Image Modeling by employing a permutation-based objective to capture the interdependencies among predicted tokens and auxiliary position information to enable the model to access a full sequence of image patches, thus reducing the discrepancy between pre-training and fine-tuning. Moreover, we introduce the $k$-CLIP tokenizer that can densely capture the semantic information of the visual input by leveraging discretized CLIP features as visual tokens. \rev{Experimental results demonstrate that our approach achieves improved fine-tuning performance on image classification and semantic segmentation, outperforming direct competitors under the same experimental setting.}

%% file: sections/supplementary.tex
In the following, we offer qualitative visualizations of the codebook associated with the proposed \mbox{$k$-CLIP} tokenizer. Furthermore, we provide additional implementation details of the proposed methodology.

\section{Additional Codebook Visualization}
\label{app:codebook}

\begin{figure*}[t]
    \begin{center}
    \includegraphics[width=\textwidth]{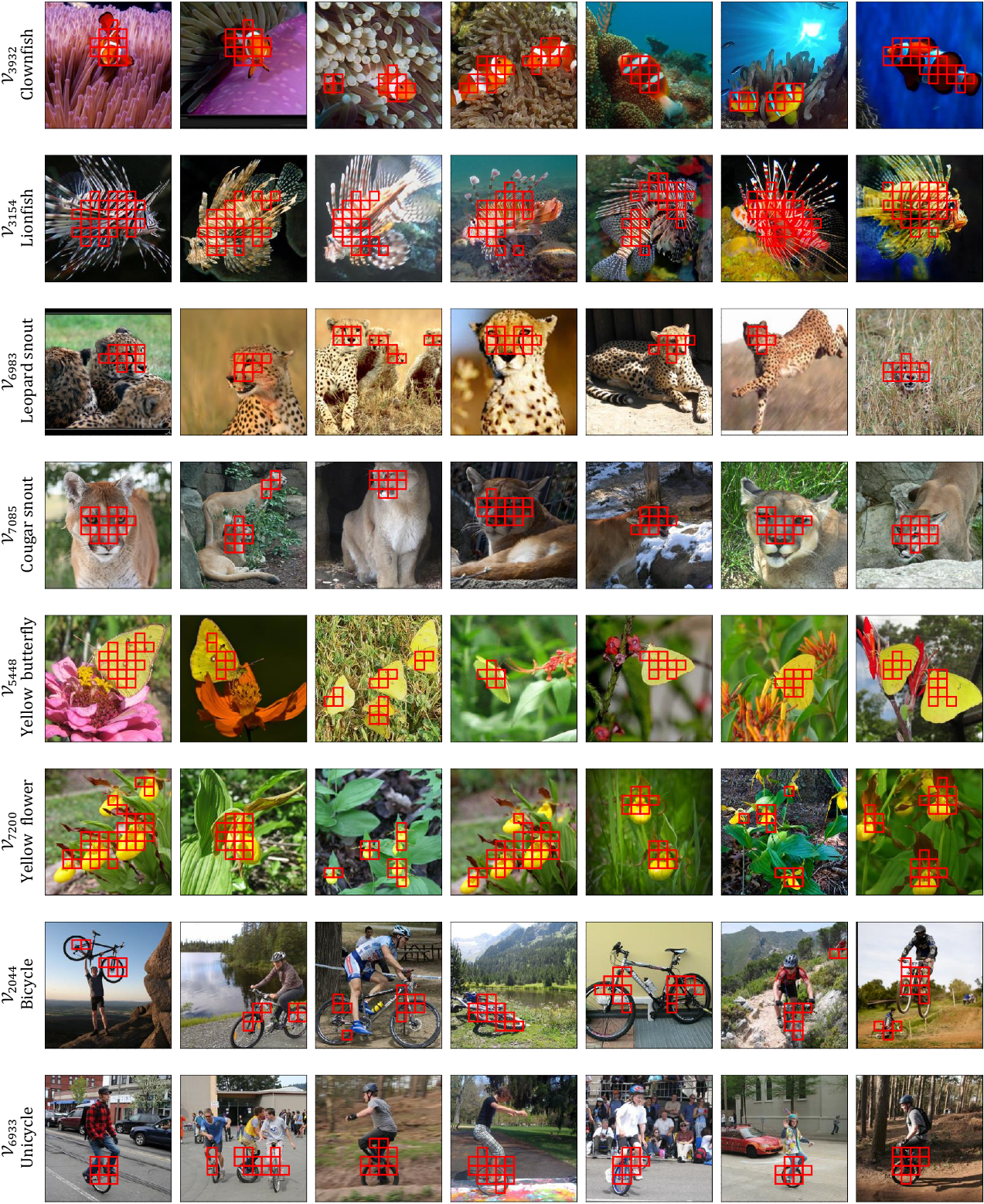}
    \end{center}
    \vspace{-.2cm}
    \caption{Visualization of image patches corresponding to the discrete tokens contained within our \mbox{\mbox{$k$-CLIP}}\ codebook. The codebook exhibits a high degree of semantic density and coherency, \ie, semantically similar image patches are consistently linked with the same discrete token in the visual codebook. Corresponding image patches are marked in \textcolor{red}{red rectangle}.}
    \label{fig:supplementary_positive}
\end{figure*}

In Fig.~\ref{fig:supplementary_positive} we report a supplementary compilation of examples previously introduced in Sec.~\ref{sec:tokenizer}. Within this collection, distinct visual tokens successfully represent semantic concepts. Notably, the pairs of \texttt{Clownfish} and \texttt{Lionfish}, \texttt{Leopard snout} and \texttt{Cougar snout}, \texttt{Yellow butterfly} and \texttt{Yellow flower}, as well as \texttt{Bicycle} and \texttt{Unicycle}, further validate the discriminative nature of visual tokens, even for highly similar categories.

\begin{figure*}[t]
    \begin{center}
    \includegraphics[width=\textwidth]{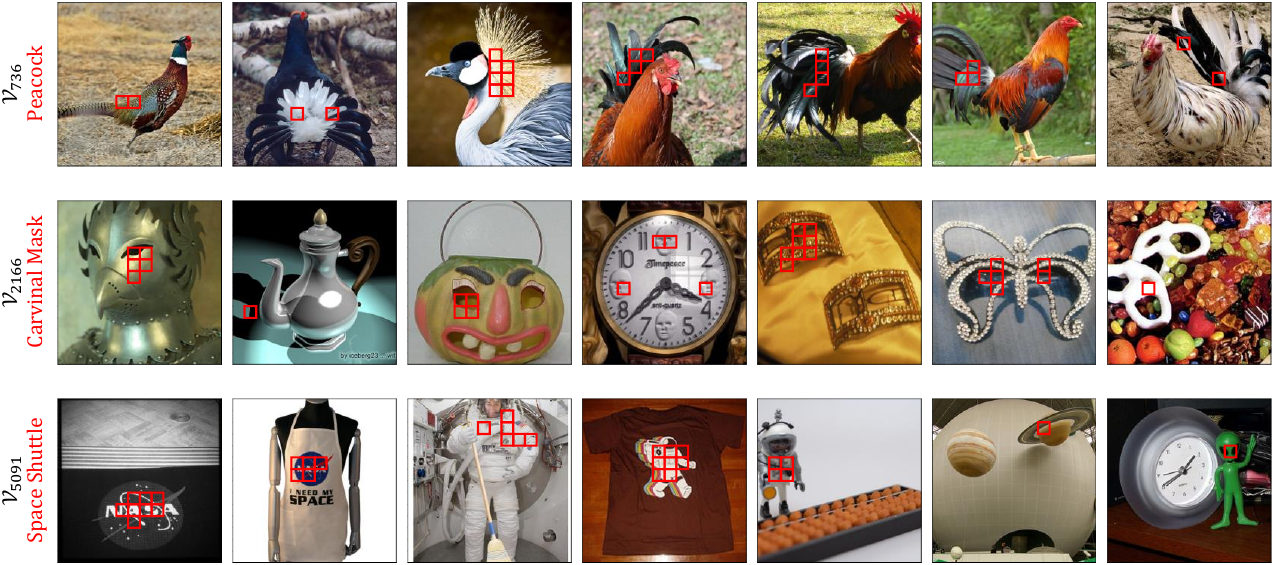}
    \end{center}
    \vspace{-.2cm}
    \caption{Visualization of image patches mismatching the semantic concept associated with the discrete tokens contained within our \mbox{\mbox{$k$-CLIP}} codebook. Corresponding image patches are marked in \textcolor{red}{red rectangle}.}
    \label{fig:supplementary_negative}
\end{figure*}

In Fig.~\ref{fig:supplementary_negative}, we also show some examples of image patches that mismatch the semantic concepts associated with the discrete tokens of the codebook. Indeed, we observe some incongruities that may be linked to the resemblance of patches to other semantic concepts. For instance, in the first row of the figure, some image patches are identified with the \texttt{Peacock} visual tokens while the bird species are different. The sumptuous tails of these birds can be visually associated with the characteristic features of a peacock. Similarly, in the second row of the image, the silver and gold perforations, which are associated with medieval headgear or adornments, may be perceived as comparable to the surface of a carnival mask.
Furthermore, within the third row of the figure, it is worth noting that several images do not precisely match the semantically correct concept of \texttt{Space Shuttle}. Nevertheless, these images demonstrate a semantic affinity with the broader, higher-level concept of \enquote{Space}. The images contain a variety of visual elements, including the NASA logo, astronauts, planets, and aliens. While the specific content of these images may diverge from the targeted concept, the overall semantic theme they evoke can be seen as related to the broader concept of space exploration and travel. These observations highlight our \mbox{$k$-CLIP} tokenizer enables the identification and classification of complex visual concepts.

\input{tables/table_hyp_pre}

\input{tables/table_hyp_fine}

\input{tables/table_hyp_seg}

\section{Additional Implementation Details}
\label{app:pretrain}

\tinytit{Hyperparameters for pre-training} In Table~\ref{tab:hyp_pre}, we schematize the experimental settings adopted during the pre-training phase.


\tit{Hyperparameters for fine-tuning}
The complete experimental configuration for fine-tuning our classification models is outlined in Table~\ref{tab:hyp_fine}.


\tit{Hyperparameters for linear probe}
Linear probing has been a widely considered proxy for assessing the effectiveness of self-supervised pre-training models. In accordance with the approach outlined in~\citep{bao2022beit}, we train a linear classifier head over the image-level representation output produced by the frozen pre-trained backbone. We use the class labels of the images to train the aforementioned classifier head. 
We train for 50 epochs using a batch size of 1024, AdamW~\citep{DBLP:conf/iclr/LoshchilovH19} as optimizer, and a learning rate of $4e{-3}$ with cosine decay. The weight decay is set to $1e{-4}$. Our pre-training augmentation strategy incorporates random resizing of crops, horizontal flipping during training, and central crops during evaluation.

\tit{\rev{Hyperparameters for semantic segmentation}}
\rev{The hyperparameters adopted for fine-tuning \ours on the ADE20K~\citep{zhou2017scene,zhou2019semantic} dataset for semantic segmentation tasks are outlined in Table~\ref{tab:hyp_seg}.}

%% file: tables/table_hyp_pre.tex
\begin{table}[t]
\centering
\setlength{\tabcolsep}{.4em}
\resizebox{0.9\linewidth}{!}{
\begin{tabular}{l|ccc}
\toprule
\bf{Hyperparameter} & \bf{Tiny Size} & \bf{Small Size} & \bf{Base Size} \\
\midrule
Hidden size & 192 & 384 & 768 \\
FFN inner hidden size & 768 & 1536 & 3072 \\
Attention heads & 3 & 6 & 12 \\
Layers & \multicolumn{3}{c}{12}  \\
Attention head size & \multicolumn{3}{c}{64} \\
Layer scale & \multicolumn{3}{c}{0.1} \\
Patch size & \multicolumn{3}{c}{$16 \times 16$} \\
\midrule
Training epochs & \multicolumn{3}{c}{300} \\
Batch size & \multicolumn{3}{c}{2048} \\
Optimizer & \multicolumn{3}{c}{AdamW}  \\
Adam $\epsilon$ & \multicolumn{3}{c}{1e-8} \\
Adam $\beta$ & \multicolumn{3}{c}{(0.9, 0.999)} \\
Peak learning rate & \multicolumn{3}{c}{1.5e-3} \\
Minimal learning rate & \multicolumn{3}{c}{1e-5} \\
Warmup learning rate & \multicolumn{3}{c}{1e-6} \\
Learning rate schedule & \multicolumn{3}{c}{Cosine} \\
Warmup epochs & \multicolumn{3}{c}{10} \\
\midrule
Gradient clipping & \multicolumn{3}{c}{3.0} \\
Dropout & \multicolumn{3}{c}{\xmark} \\
Stoch. depth & \multicolumn{3}{c}{0.1} \\
Weight decay & \multicolumn{3}{c}{0.05} \\
\midrule
Input resolution & \multicolumn{3}{c}{$224 \times 224$} \\
Color jitter & \multicolumn{3}{c}{0.4} \\
\bottomrule
\end{tabular}
}
\caption{Hyperparameters for pre-training on ImageNet-1k~\citep{deng2009imagenet}.}
\label{tab:hyp_pre}
\end{table}

%% file: tables/table_hyp_fine.tex
\begin{table}[th!]
\centering
\setlength{\tabcolsep}{.4em}
\resizebox{1.\linewidth}{!}{
\begin{tabular}{l|ccc}
\toprule
\bf{Hyperparameter} & \bf{Tiny Size} & \bf{Small Size} & \bf{Base Size} \\
\midrule
Peak learning rate & 2.5e-4 & 5e-3 & 5e-4 \\
Fine-tuning epochs & 300 & 200 & 100 \\
Layer-wise learning rate decay & \xmark & 0.65 & 0.65 \\
Batch size & \multicolumn{3}{c}{1024} \\
Warmup epochs & \multicolumn{3}{c}{20} \\
Optimizer & \multicolumn{3}{c}{AdamW}  \\
Adam $\epsilon$ & \multicolumn{3}{c}{1e-8}  \\
Adam $\beta$ & \multicolumn{3}{c}{(0.9, 0.999)} \\
Minimal learning rate & \multicolumn{3}{c}{1e-6} \\
Warmup learning rate & \multicolumn{3}{c}{1e-6} \\
Learning rate schedule & \multicolumn{3}{c}{Cosine} \\
\midrule
Weight decay & \multicolumn{3}{c}{0.05} \\
Label smoothing $\varepsilon$ & \multicolumn{3}{c}{0.1}     \\
Stoch. depth & \multicolumn{3}{c}{0.1} \\
Dropout & \multicolumn{3}{c}{\xmark} \\
Gradient clipping & \multicolumn{3}{c}{\xmark} \\
\midrule
Erasing prob.  & \multicolumn{3}{c}{0.25} \\
Input resolution & \multicolumn{3}{c}{$224 \times 224$} \\
Rand Augment  & \multicolumn{3}{c}{9/0.5} \\
Mixup prob.  & \multicolumn{3}{c}{0.8}     \\
Cutmix prob.   & \multicolumn{3}{c}{1.0}    \\
\bottomrule
\end{tabular}
}
\caption{Hyperparameters for fine-tuning on ImageNet-1k~\citep{deng2009imagenet}.}
\label{tab:hyp_fine}
\end{table}

%% file: tables/table_hyp_seg.tex
\begin{table}[th!]
\centering
\setlength{\tabcolsep}{.4em}
\resizebox{0.85\linewidth}{!}{
\begin{tabular}{>{\color{black}}l|>{\color{black}}c}
\toprule
\bf{Hyperparameter} & \bf{Tiny\,/\,Small\,/\,Base Size} \\
\midrule
Input resolution & $512 \times 512$ \\
\midrule
Peak learning rate & 5e-5  \\
Fine-tuning steps & 160k \\
Batch size & 16  \\
Optimizer & AdamW  \\ 
Adam $\epsilon$ & 1e-8  \\
Adam $\beta$ & (0.9, 0.999)  \\
Layer-wise learning rate decay & 0.75  \\
Minimal learning rate & 0  \\
Warmup learning rate & 5e-11  \\
Learning rate schedule & Linear  \\
Warmup steps & 1500 \\
\midrule
Weight decay & 0.05 \\
Stochastic depth & 0.15  \\
Dropout & \xmark  \\
\bottomrule
\end{tabular}
}
\caption{\rev{Hyperparameters for fine-tuning \ours on ADE20K~\citep{zhou2017scene,zhou2019semantic} for semantic segmentation tasks.}}
\label{tab:hyp_seg}
\end{table}

%% file: main.bbl
\begin{thebibliography}{46}
\expandafter\ifx\csname natexlab\endcsname\relax\def\natexlab#1{#1}\fi
\providecommand{\url}[1]{\texttt{#1}}
\providecommand{\href}[2]{#2}
\providecommand{\path}[1]{#1}
\providecommand{\DOIprefix}{doi:}
\providecommand{\ArXivprefix}{arXiv:}
\providecommand{\URLprefix}{URL: }
\providecommand{\Pubmedprefix}{pmid:}
\providecommand{\doi}[1]{\href{http://dx.doi.org/#1}{\path{#1}}}
\providecommand{\Pubmed}[1]{\href{pmid:#1}{\path{#1}}}
\providecommand{\bibinfo}[2]{#2}
\ifx\xfnm\relax \def\xfnm[#1]{\unskip,\space#1}\fi
\bibitem[{Bao et~al.(2022)Bao, Dong, Piao and Wei}]{bao2022beit}
\bibinfo{author}{Bao, H.}, \bibinfo{author}{Dong, L.}, \bibinfo{author}{Piao, S.}, \bibinfo{author}{Wei, F.}, \bibinfo{year}{2022}.
\newblock \bibinfo{title}{{BEiT: BERT pre-training of image Transformers}}, in: \bibinfo{booktitle}{Proceedings of the International Conference on Learning Representations}.
\bibitem[{Bossard et~al.(2014)Bossard, Guillaumin and Van~Gool}]{bossard2014food}
\bibinfo{author}{Bossard, L.}, \bibinfo{author}{Guillaumin, M.}, \bibinfo{author}{Van~Gool, L.}, \bibinfo{year}{2014}.
\newblock \bibinfo{title}{Food-101--mining discriminative components with random forests}, in: \bibinfo{booktitle}{Proceedings of the European Conference on Computer Vision}.
\bibitem[{Brown et~al.(2020)Brown, Mann, Ryder, Subbiah, Kaplan, Dhariwal, Neelakantan, Shyam, Sastry, Askell et~al.}]{brown2020language}
\bibinfo{author}{Brown, T.}, \bibinfo{author}{Mann, B.}, \bibinfo{author}{Ryder, N.}, \bibinfo{author}{Subbiah, M.}, \bibinfo{author}{Kaplan, J.D.}, \bibinfo{author}{Dhariwal, P.}, \bibinfo{author}{Neelakantan, A.}, \bibinfo{author}{Shyam, P.}, \bibinfo{author}{Sastry, G.}, \bibinfo{author}{Askell, A.}, et~al., \bibinfo{year}{2020}.
\newblock \bibinfo{title}{{Language Models are Few-Shot Learners}}, in: \bibinfo{booktitle}{Advances in Neural Information Processing Systems}.
\bibitem[{Caron et~al.(2021)Caron, Touvron, Misra, J{\'e}gou, Mairal, Bojanowski and Joulin}]{caron2021emerging}
\bibinfo{author}{Caron, M.}, \bibinfo{author}{Touvron, H.}, \bibinfo{author}{Misra, I.}, \bibinfo{author}{J{\'e}gou, H.}, \bibinfo{author}{Mairal, J.}, \bibinfo{author}{Bojanowski, P.}, \bibinfo{author}{Joulin, A.}, \bibinfo{year}{2021}.
\newblock \bibinfo{title}{Emerging properties in self-supervised vision transformers}, in: \bibinfo{booktitle}{Proceedings of the IEEE/CVF International Conference on Computer Vision}.
\bibitem[{Chen et~al.(2020a)Chen, Radford, Child, Wu, Jun, Luan and Sutskever}]{chen2020generative}
\bibinfo{author}{Chen, M.}, \bibinfo{author}{Radford, A.}, \bibinfo{author}{Child, R.}, \bibinfo{author}{Wu, J.}, \bibinfo{author}{Jun, H.}, \bibinfo{author}{Luan, D.}, \bibinfo{author}{Sutskever, I.}, \bibinfo{year}{2020}a.
\newblock \bibinfo{title}{{Generative Pretraining From Pixels}}, in: \bibinfo{booktitle}{Proceedings of the International Conference on Machine Learning}.
\bibitem[{Chen et~al.(2020b)Chen, Kornblith, Norouzi and Hinton}]{chen2020simple}
\bibinfo{author}{Chen, T.}, \bibinfo{author}{Kornblith, S.}, \bibinfo{author}{Norouzi, M.}, \bibinfo{author}{Hinton, G.}, \bibinfo{year}{2020}b.
\newblock \bibinfo{title}{A simple framework for contrastive learning of visual representations}, in: \bibinfo{booktitle}{Proceedings of the International Conference on Machine Learning}.
\bibitem[{Chen et~al.(2023)Chen, Ding, Wang, Xin, Mo, Wang, Han, Luo, Zeng and Wang}]{chen2022context}
\bibinfo{author}{Chen, X.}, \bibinfo{author}{Ding, M.}, \bibinfo{author}{Wang, X.}, \bibinfo{author}{Xin, Y.}, \bibinfo{author}{Mo, S.}, \bibinfo{author}{Wang, Y.}, \bibinfo{author}{Han, S.}, \bibinfo{author}{Luo, P.}, \bibinfo{author}{Zeng, G.}, \bibinfo{author}{Wang, J.}, \bibinfo{year}{2023}.
\newblock \bibinfo{title}{{Context Autoencoder for Self-Supervised Representation Learning}}.
\newblock \bibinfo{journal}{International Journal of Computer Vision} .
\bibitem[{Chen and He(2021)}]{chen2021exploring}
\bibinfo{author}{Chen, X.}, \bibinfo{author}{He, K.}, \bibinfo{year}{2021}.
\newblock \bibinfo{title}{Exploring simple siamese representation learning}, in: \bibinfo{booktitle}{Proceedings of the IEEE/CVF Conference on Computer Vision and Pattern Recognition}.
\bibitem[{Chen et~al.(2021)Chen, Xie and He}]{chen2021empirical}
\bibinfo{author}{Chen, X.}, \bibinfo{author}{Xie, S.}, \bibinfo{author}{He, K.}, \bibinfo{year}{2021}.
\newblock \bibinfo{title}{{An Empirical Study of Training Self-Supervised Vision Transformers}}, in: \bibinfo{booktitle}{Proceedings of the IEEE/CVF International Conference on Computer Vision}.
\bibitem[{Deng et~al.(2009)Deng, Dong, Socher, Li, Li and Fei-Fei}]{deng2009imagenet}
\bibinfo{author}{Deng, J.}, \bibinfo{author}{Dong, W.}, \bibinfo{author}{Socher, R.}, \bibinfo{author}{Li, L.J.}, \bibinfo{author}{Li, K.}, \bibinfo{author}{Fei-Fei, L.}, \bibinfo{year}{2009}.
\newblock \bibinfo{title}{{ImageNet: A large-scale hierarchical image database}}, in: \bibinfo{booktitle}{Proceedings of the IEEE/CVF Conference on Computer Vision and Pattern Recognition}.
\bibitem[{Devlin et~al.(2019)Devlin, Chang, Lee and Toutanova}]{devlin2019bert}
\bibinfo{author}{Devlin, J.}, \bibinfo{author}{Chang, M.W.}, \bibinfo{author}{Lee, K.}, \bibinfo{author}{Toutanova, K.}, \bibinfo{year}{2019}.
\newblock \bibinfo{title}{{BERT: Pre-training of Deep Bidirectional Transformers for Language Understanding}}, in: \bibinfo{booktitle}{Proceedings of the Annual Conference of the North American Chapter of the Association for Computational Linguistics}.
\bibitem[{Doersch et~al.(2015)Doersch, Gupta and Efros}]{doersch2015unsupervised}
\bibinfo{author}{Doersch, C.}, \bibinfo{author}{Gupta, A.}, \bibinfo{author}{Efros, A.A.}, \bibinfo{year}{2015}.
\newblock \bibinfo{title}{{Unsupervised Visual Representation Learning by Context Prediction}}, in: \bibinfo{booktitle}{Proceedings of the IEEE/CVF International Conference on Computer Vision}.
\bibitem[{Dong et~al.(2023)Dong, Bao, Zhang, Chen, Zhang, Yuan, Chen, Wen and Yu}]{dong2023peco}
\bibinfo{author}{Dong, X.}, \bibinfo{author}{Bao, J.}, \bibinfo{author}{Zhang, T.}, \bibinfo{author}{Chen, D.}, \bibinfo{author}{Zhang, W.}, \bibinfo{author}{Yuan, L.}, \bibinfo{author}{Chen, D.}, \bibinfo{author}{Wen, F.}, \bibinfo{author}{Yu, N.}, \bibinfo{year}{2023}.
\newblock \bibinfo{title}{{PeCo: Perceptual Codebook for BERT Pre-training of Vision Transformers}}, in: \bibinfo{booktitle}{Proceedings of the AAAI Conference on Artificial Intelligence}.
\bibitem[{Dosovitskiy et~al.(2021)Dosovitskiy, Beyer, Kolesnikov, Weissenborn, Zhai, Unterthiner, Dehghani, Minderer, Heigold, Gelly, Uszkoreit and Houlsby}]{dosovitskiy2021image}
\bibinfo{author}{Dosovitskiy, A.}, \bibinfo{author}{Beyer, L.}, \bibinfo{author}{Kolesnikov, A.}, \bibinfo{author}{Weissenborn, D.}, \bibinfo{author}{Zhai, X.}, \bibinfo{author}{Unterthiner, T.}, \bibinfo{author}{Dehghani, M.}, \bibinfo{author}{Minderer, M.}, \bibinfo{author}{Heigold, G.}, \bibinfo{author}{Gelly, S.}, \bibinfo{author}{Uszkoreit, J.}, \bibinfo{author}{Houlsby, N.}, \bibinfo{year}{2021}.
\newblock \bibinfo{title}{{An Image is Worth 16x16 Words: Transformers for Image Recognition at Scale}}, in: \bibinfo{booktitle}{Proceedings of the International Conference on Learning Representations}.
\bibitem[{El-Nouby et~al.(2021)El-Nouby, Izacard, Touvron, Laptev, Jegou and Grave}]{el2021large}
\bibinfo{author}{El-Nouby, A.}, \bibinfo{author}{Izacard, G.}, \bibinfo{author}{Touvron, H.}, \bibinfo{author}{Laptev, I.}, \bibinfo{author}{Jegou, H.}, \bibinfo{author}{Grave, E.}, \bibinfo{year}{2021}.
\newblock \bibinfo{title}{{Are Large-scale Datasets Necessary for Self-Supervised Pre-training?}}
\newblock \bibinfo{journal}{arXiv preprint arXiv:2112.10740} .
\bibitem[{Fang et~al.(2023)Fang, Dong, Bao, Wang and Wei}]{fang2022corrupted}
\bibinfo{author}{Fang, Y.}, \bibinfo{author}{Dong, L.}, \bibinfo{author}{Bao, H.}, \bibinfo{author}{Wang, X.}, \bibinfo{author}{Wei, F.}, \bibinfo{year}{2023}.
\newblock \bibinfo{title}{Corrupted image modeling for self-supervised visual pre-training}, in: \bibinfo{booktitle}{Proceedings of the International Conference on Learning Representations}.
\bibitem[{Gidaris et~al.(2018)Gidaris, Singh and Komodakis}]{gidaris2018unsupervised}
\bibinfo{author}{Gidaris, S.}, \bibinfo{author}{Singh, P.}, \bibinfo{author}{Komodakis, N.}, \bibinfo{year}{2018}.
\newblock \bibinfo{title}{Unsupervised representation learning by predicting image rotations}, in: \bibinfo{booktitle}{Proceedings of the International Conference on Learning Representations}.
\bibitem[{Grill et~al.(2020)Grill, Strub, Altch{\'e}, Tallec, Richemond, Buchatskaya, Doersch, Avila~Pires, Guo, Gheshlaghi~Azar, Piot, Kavukcuoglu, Munos and Valko}]{grill2020bootstrap}
\bibinfo{author}{Grill, J.B.}, \bibinfo{author}{Strub, F.}, \bibinfo{author}{Altch{\'e}, F.}, \bibinfo{author}{Tallec, C.}, \bibinfo{author}{Richemond, P.}, \bibinfo{author}{Buchatskaya, E.}, \bibinfo{author}{Doersch, C.}, \bibinfo{author}{Avila~Pires, B.}, \bibinfo{author}{Guo, Z.}, \bibinfo{author}{Gheshlaghi~Azar, M.}, \bibinfo{author}{Piot, B.}, \bibinfo{author}{Kavukcuoglu, K.}, \bibinfo{author}{Munos, R.}, \bibinfo{author}{Valko, M.}, \bibinfo{year}{2020}.
\newblock \bibinfo{title}{Bootstrap your own latent-a new approach to self-supervised learning}, in: \bibinfo{booktitle}{Advances in Neural Information Processing Systems}.
\bibitem[{He et~al.(2022)He, Chen, Xie, Li, Doll{\'a}r and Girshick}]{he2022masked}
\bibinfo{author}{He, K.}, \bibinfo{author}{Chen, X.}, \bibinfo{author}{Xie, S.}, \bibinfo{author}{Li, Y.}, \bibinfo{author}{Doll{\'a}r, P.}, \bibinfo{author}{Girshick, R.}, \bibinfo{year}{2022}.
\newblock \bibinfo{title}{{Masked Autoencoders Are Scalable Vision Learners}}, in: \bibinfo{booktitle}{Proceedings of the IEEE/CVF Conference on Computer Vision and Pattern Recognition}.
\bibitem[{He et~al.(2020)He, Fan, Wu, Xie and Girshick}]{he2020momentum}
\bibinfo{author}{He, K.}, \bibinfo{author}{Fan, H.}, \bibinfo{author}{Wu, Y.}, \bibinfo{author}{Xie, S.}, \bibinfo{author}{Girshick, R.}, \bibinfo{year}{2020}.
\newblock \bibinfo{title}{Momentum contrast for unsupervised visual representation learning}, in: \bibinfo{booktitle}{Proceedings of the IEEE/CVF Conference on Computer Vision and Pattern Recognition}.
\bibitem[{Hou et~al.(2022)Hou, Sun, Chen, Xie and Kung}]{MILAN2022}
\bibinfo{author}{Hou, Z.}, \bibinfo{author}{Sun, F.}, \bibinfo{author}{Chen, Y.K.}, \bibinfo{author}{Xie, Y.}, \bibinfo{author}{Kung, S.Y.}, \bibinfo{year}{2022}.
\newblock \bibinfo{title}{{MILAN: Masked Image Pretraining on Language Assisted Representation}}.
\newblock \bibinfo{journal}{arXiv preprint arXiv:2208.06049} .
\bibitem[{Huang et~al.(2023)Huang, Jin, Lu, Hou, Cheng, Fu, Shen and Feng}]{huang2022contrastive}
\bibinfo{author}{Huang, Z.}, \bibinfo{author}{Jin, X.}, \bibinfo{author}{Lu, C.}, \bibinfo{author}{Hou, Q.}, \bibinfo{author}{Cheng, M.M.}, \bibinfo{author}{Fu, D.}, \bibinfo{author}{Shen, X.}, \bibinfo{author}{Feng, J.}, \bibinfo{year}{2023}.
\newblock \bibinfo{title}{{Contrastive Masked Autoencoders are Stronger Vision Learners}}.
\newblock \bibinfo{journal}{IEEE Transactions on Pattern Analysis and Machine Intelligence} .
\bibitem[{Krause et~al.(2013)Krause, Stark, Deng and Fei-Fei}]{krause20133d}
\bibinfo{author}{Krause, J.}, \bibinfo{author}{Stark, M.}, \bibinfo{author}{Deng, J.}, \bibinfo{author}{Fei-Fei, L.}, \bibinfo{year}{2013}.
\newblock \bibinfo{title}{{3D Object Representations for Fine-Grained Categorization}}, in: \bibinfo{booktitle}{Proceedings of the IEEE/CVF International Conference on Computer Vision}.
\bibitem[{Liu et~al.(2019)Liu, Ott, Goyal, Du, Joshi, Chen, Levy, Lewis, Zettlemoyer and Stoyanov}]{liu2019roberta}
\bibinfo{author}{Liu, Y.}, \bibinfo{author}{Ott, M.}, \bibinfo{author}{Goyal, N.}, \bibinfo{author}{Du, J.}, \bibinfo{author}{Joshi, M.}, \bibinfo{author}{Chen, D.}, \bibinfo{author}{Levy, O.}, \bibinfo{author}{Lewis, M.}, \bibinfo{author}{Zettlemoyer, L.}, \bibinfo{author}{Stoyanov, V.}, \bibinfo{year}{2019}.
\newblock \bibinfo{title}{{RoBERTa: A Robustly Optimized BERT Pretraining Approach}}.
\newblock \bibinfo{journal}{arXiv preprint arXiv:1907.11692} .
\bibitem[{Liu et~al.(2022)Liu, Hu, Lin, Yao, Xie, Wei, Ning, Cao, Zhang, Dong et~al.}]{liu2022swin}
\bibinfo{author}{Liu, Z.}, \bibinfo{author}{Hu, H.}, \bibinfo{author}{Lin, Y.}, \bibinfo{author}{Yao, Z.}, \bibinfo{author}{Xie, Z.}, \bibinfo{author}{Wei, Y.}, \bibinfo{author}{Ning, J.}, \bibinfo{author}{Cao, Y.}, \bibinfo{author}{Zhang, Z.}, \bibinfo{author}{Dong, L.}, et~al., \bibinfo{year}{2022}.
\newblock \bibinfo{title}{{Swin Transformer V2: Scaling Up Capacity and Resolution}}, in: \bibinfo{booktitle}{Proceedings of the IEEE/CVF Conference on Computer Vision and Pattern Recognition}.
\bibitem[{Loshchilov and Hutter(2019)}]{DBLP:conf/iclr/LoshchilovH19}
\bibinfo{author}{Loshchilov, I.}, \bibinfo{author}{Hutter, F.}, \bibinfo{year}{2019}.
\newblock \bibinfo{title}{Decoupled weight decay regularization}, in: \bibinfo{booktitle}{Proceedings of the International Conference on Learning Representations}.
\bibitem[{Maji et~al.(2013)Maji, Kannala, Rahtu, Blaschko and Vedaldi}]{maji13fine-grained}
\bibinfo{author}{Maji, S.}, \bibinfo{author}{Kannala, J.}, \bibinfo{author}{Rahtu, E.}, \bibinfo{author}{Blaschko, M.}, \bibinfo{author}{Vedaldi, A.}, \bibinfo{year}{2013}.
\newblock \bibinfo{title}{Fine-Grained Visual Classification of Aircraft}.
\newblock \bibinfo{type}{Technical Report}.
\bibitem[{Noroozi and Favaro(2016)}]{noroozi2016unsupervised}
\bibinfo{author}{Noroozi, M.}, \bibinfo{author}{Favaro, P.}, \bibinfo{year}{2016}.
\newblock \bibinfo{title}{{Unsupervised Learning of Visual Representations by Solving Jigsaw Puzzles}}, in: \bibinfo{booktitle}{Proceedings of the European Conference on Computer Vision}.
\bibitem[{Oord et~al.(2018)Oord, Li and Vinyals}]{oord2018representation}
\bibinfo{author}{Oord, A.v.d.}, \bibinfo{author}{Li, Y.}, \bibinfo{author}{Vinyals, O.}, \bibinfo{year}{2018}.
\newblock \bibinfo{title}{Representation learning with contrastive predictive coding}.
\newblock \bibinfo{journal}{arXiv preprint arXiv:1807.03748} .
\bibitem[{Peng et~al.(2022)Peng, Dong, Bao, Ye and Wei}]{peng2022beit}
\bibinfo{author}{Peng, Z.}, \bibinfo{author}{Dong, L.}, \bibinfo{author}{Bao, H.}, \bibinfo{author}{Ye, Q.}, \bibinfo{author}{Wei, F.}, \bibinfo{year}{2022}.
\newblock \bibinfo{title}{{BEiT v2: Masked Image Modeling with Vector-Quantized Visual Tokenizers}}.
\newblock \bibinfo{journal}{arXiv preprint arXiv:2208.06366} .
\bibitem[{Radford et~al.(2021)Radford, Kim, Hallacy, Ramesh, Goh, Agarwal, Sastry, Askell, Mishkin, Clark, Krueger and Sutskever}]{radford2021learning}
\bibinfo{author}{Radford, A.}, \bibinfo{author}{Kim, J.W.}, \bibinfo{author}{Hallacy, C.}, \bibinfo{author}{Ramesh, A.}, \bibinfo{author}{Goh, G.}, \bibinfo{author}{Agarwal, S.}, \bibinfo{author}{Sastry, G.}, \bibinfo{author}{Askell, A.}, \bibinfo{author}{Mishkin, P.}, \bibinfo{author}{Clark, J.}, \bibinfo{author}{Krueger, G.}, \bibinfo{author}{Sutskever, I.}, \bibinfo{year}{2021}.
\newblock \bibinfo{title}{Learning transferable visual models from natural language supervision}, in: \bibinfo{booktitle}{Proceedings of the International Conference on Machine Learning}.
\bibitem[{Ramesh et~al.(2021)Ramesh, Pavlov, Goh, Gray, Voss, Radford, Chen and Sutskever}]{ramesh2021zero}
\bibinfo{author}{Ramesh, A.}, \bibinfo{author}{Pavlov, M.}, \bibinfo{author}{Goh, G.}, \bibinfo{author}{Gray, S.}, \bibinfo{author}{Voss, C.}, \bibinfo{author}{Radford, A.}, \bibinfo{author}{Chen, M.}, \bibinfo{author}{Sutskever, I.}, \bibinfo{year}{2021}.
\newblock \bibinfo{title}{{Zero-Shot Text-to-Image Generation}}, in: \bibinfo{booktitle}{Proceedings of the International Conference on Machine Learning}.
\bibitem[{Song et~al.(2020)Song, Tan, Qin, Lu and Liu}]{song2020mpnet}
\bibinfo{author}{Song, K.}, \bibinfo{author}{Tan, X.}, \bibinfo{author}{Qin, T.}, \bibinfo{author}{Lu, J.}, \bibinfo{author}{Liu, T.Y.}, \bibinfo{year}{2020}.
\newblock \bibinfo{title}{{MPNet: Masked and Permuted Pre-training for Language Understanding}}, in: \bibinfo{booktitle}{Advances in Neural Information Processing Systems}.
\bibitem[{Tian et~al.(2022)Tian, Xie, Fang, Shi, Peng, Zhang, Jiao, Tian and Ye}]{tian2022beyond}
\bibinfo{author}{Tian, Y.}, \bibinfo{author}{Xie, L.}, \bibinfo{author}{Fang, J.}, \bibinfo{author}{Shi, M.}, \bibinfo{author}{Peng, J.}, \bibinfo{author}{Zhang, X.}, \bibinfo{author}{Jiao, J.}, \bibinfo{author}{Tian, Q.}, \bibinfo{author}{Ye, Q.}, \bibinfo{year}{2022}.
\newblock \bibinfo{title}{Beyond masking: Demystifying token-based pre-training for vision transformers}.
\newblock \bibinfo{journal}{arXiv preprint arXiv:2203.14313} .
\bibitem[{Tian et~al.(2023)Tian, Xie, Wang, Wei, Zhang, Jiao, Wang, Tian and Ye}]{tian2023integrally}
\bibinfo{author}{Tian, Y.}, \bibinfo{author}{Xie, L.}, \bibinfo{author}{Wang, Z.}, \bibinfo{author}{Wei, L.}, \bibinfo{author}{Zhang, X.}, \bibinfo{author}{Jiao, J.}, \bibinfo{author}{Wang, Y.}, \bibinfo{author}{Tian, Q.}, \bibinfo{author}{Ye, Q.}, \bibinfo{year}{2023}.
\newblock \bibinfo{title}{Integrally pre-trained transformer pyramid networks}, in: \bibinfo{booktitle}{Proceedings of the IEEE/CVF Conference on Computer Vision and Pattern Recognition}.
\bibitem[{Wang and Gupta(2015)}]{wang2015unsupervised}
\bibinfo{author}{Wang, X.}, \bibinfo{author}{Gupta, A.}, \bibinfo{year}{2015}.
\newblock \bibinfo{title}{Unsupervised learning of visual representations using videos}, in: \bibinfo{booktitle}{Proceedings of the IEEE/CVF International Conference on Computer Vision}.
\bibitem[{Wei et~al.(2022a)Wei, Fan, Xie, Wu, Yuille and Feichtenhofer}]{wei2022masked}
\bibinfo{author}{Wei, C.}, \bibinfo{author}{Fan, H.}, \bibinfo{author}{Xie, S.}, \bibinfo{author}{Wu, C.Y.}, \bibinfo{author}{Yuille, A.}, \bibinfo{author}{Feichtenhofer, C.}, \bibinfo{year}{2022}a.
\newblock \bibinfo{title}{Masked feature prediction for self-supervised visual pre-training}, in: \bibinfo{booktitle}{Proceedings of the IEEE/CVF Conference on Computer Vision and Pattern Recognition}.
\bibitem[{Wei et~al.(2022b)Wei, Xie, Zhou, Li and Tian}]{wei2022mvp}
\bibinfo{author}{Wei, L.}, \bibinfo{author}{Xie, L.}, \bibinfo{author}{Zhou, W.}, \bibinfo{author}{Li, H.}, \bibinfo{author}{Tian, Q.}, \bibinfo{year}{2022}b.
\newblock \bibinfo{title}{{MVP: Multimodality-guided Visual Pre-training}}, in: \bibinfo{booktitle}{Proceedings of the European Conference on Computer Vision}.
\bibitem[{Xiao et~al.(2018)Xiao, Liu, Zhou, Jiang and Sun}]{xiao2018unified}
\bibinfo{author}{Xiao, T.}, \bibinfo{author}{Liu, Y.}, \bibinfo{author}{Zhou, B.}, \bibinfo{author}{Jiang, Y.}, \bibinfo{author}{Sun, J.}, \bibinfo{year}{2018}.
\newblock \bibinfo{title}{Unified perceptual parsing for scene understanding}, in: \bibinfo{booktitle}{Proceedings of the European Conference on Computer Vision}.
\bibitem[{Xie et~al.(2022)Xie, Zhang, Cao, Lin, Bao, Yao, Dai and Hu}]{xie2022simmim}
\bibinfo{author}{Xie, Z.}, \bibinfo{author}{Zhang, Z.}, \bibinfo{author}{Cao, Y.}, \bibinfo{author}{Lin, Y.}, \bibinfo{author}{Bao, J.}, \bibinfo{author}{Yao, Z.}, \bibinfo{author}{Dai, Q.}, \bibinfo{author}{Hu, H.}, \bibinfo{year}{2022}.
\newblock \bibinfo{title}{{SimMIM: a Simple Framework for Masked Image Modeling}}, in: \bibinfo{booktitle}{Proceedings of the IEEE/CVF Conference on Computer Vision and Pattern Recognition}.
\bibitem[{Yang et~al.(2019)Yang, Dai, Yang, Carbonell, Salakhutdinov and Le}]{yang2019xlnet}
\bibinfo{author}{Yang, Z.}, \bibinfo{author}{Dai, Z.}, \bibinfo{author}{Yang, Y.}, \bibinfo{author}{Carbonell, J.}, \bibinfo{author}{Salakhutdinov, R.R.}, \bibinfo{author}{Le, Q.V.}, \bibinfo{year}{2019}.
\newblock \bibinfo{title}{{XLNet: Generalized Autoregressive Pretraining for Language Understanding}}, in: \bibinfo{booktitle}{Advances in Neural Information Processing Systems}.
\bibitem[{Zhang et~al.(2016)Zhang, Isola and Efros}]{zhang2016colorful}
\bibinfo{author}{Zhang, R.}, \bibinfo{author}{Isola, P.}, \bibinfo{author}{Efros, A.A.}, \bibinfo{year}{2016}.
\newblock \bibinfo{title}{Colorful image colorization}, in: \bibinfo{booktitle}{Proceedings of the European Conference on Computer Vision}.
\bibitem[{Zhang et~al.(2022)Zhang, Chen, Yuan, Chen, Wang, Wang, Han, Chen, Pi, Yao, Han, Ding and Wang}]{zhang2022cae}
\bibinfo{author}{Zhang, X.}, \bibinfo{author}{Chen, J.}, \bibinfo{author}{Yuan, J.}, \bibinfo{author}{Chen, Q.}, \bibinfo{author}{Wang, J.}, \bibinfo{author}{Wang, X.}, \bibinfo{author}{Han, S.}, \bibinfo{author}{Chen, X.}, \bibinfo{author}{Pi, J.}, \bibinfo{author}{Yao, K.}, \bibinfo{author}{Han, J.}, \bibinfo{author}{Ding, E.}, \bibinfo{author}{Wang, J.}, \bibinfo{year}{2022}.
\newblock \bibinfo{title}{{CAE v2: Context Autoencoder with CLIP Target}}.
\newblock \bibinfo{journal}{arXiv preprint arXiv:2211.09799} .
\bibitem[{Zhang et~al.(2023)Zhang, Tian, Huang, Ye, Dai, Xie and Tian}]{zhang2023hivit}
\bibinfo{author}{Zhang, X.}, \bibinfo{author}{Tian, Y.}, \bibinfo{author}{Huang, W.}, \bibinfo{author}{Ye, Q.}, \bibinfo{author}{Dai, Q.}, \bibinfo{author}{Xie, L.}, \bibinfo{author}{Tian, Q.}, \bibinfo{year}{2023}.
\newblock \bibinfo{title}{Hivit: Hierarchical vision transformer meets masked image modeling}.
\newblock \bibinfo{journal}{Proceedings of the International Conference on Learning Representations} .
\bibitem[{Zhou et~al.(2017)Zhou, Zhao, Puig, Fidler, Barriuso and Torralba}]{zhou2017scene}
\bibinfo{author}{Zhou, B.}, \bibinfo{author}{Zhao, H.}, \bibinfo{author}{Puig, X.}, \bibinfo{author}{Fidler, S.}, \bibinfo{author}{Barriuso, A.}, \bibinfo{author}{Torralba, A.}, \bibinfo{year}{2017}.
\newblock \bibinfo{title}{{Scene Parsing Through ADE20K Dataset}}, in: \bibinfo{booktitle}{Proceedings of the IEEE/CVF Conference on Computer Vision and Pattern Recognition}.
\bibitem[{Zhou et~al.(2019)Zhou, Zhao, Puig, Xiao, Fidler, Barriuso and Torralba}]{zhou2019semantic}
\bibinfo{author}{Zhou, B.}, \bibinfo{author}{Zhao, H.}, \bibinfo{author}{Puig, X.}, \bibinfo{author}{Xiao, T.}, \bibinfo{author}{Fidler, S.}, \bibinfo{author}{Barriuso, A.}, \bibinfo{author}{Torralba, A.}, \bibinfo{year}{2019}.
\newblock \bibinfo{title}{{Semantic Understanding of Scenes Through the ADE20K Dataset}}.
\newblock \bibinfo{journal}{International Journal of Computer Vision} \bibinfo{volume}{127}, \bibinfo{pages}{302--321}.

\end{thebibliography}
